\documentclass[final,5p,times,twocolumn]{elsarticle}

\usepackage[usenames,dvipsnames,table]{xcolor}
\colorlet{LightRed}{White!80!Red}
\colorlet{DarkGreen}{Black!20!Green}

\newcommand{\ADD}[1]{#1}

\usepackage{booktabs}
\definecolor{tablegray}{gray}{0.93}

\usepackage{xcolor}
\usepackage{tikz}
\usetikzlibrary{bayesnet}
\usepackage{multirow}
\usepackage{tablefootnote}
\usepackage{ragged2e}
\usepackage{amssymb}
\usepackage[tableposition=top]{caption}
\usepackage{dsfont}
\usepackage{subcaption}
\usepackage{amsmath}
\usepackage{tabularx}
\usepackage{csvsimple}
\usepackage{color}
\usepackage{xcolor,colortbl}
\usepackage[font=small,labelfont=bf]{caption}
\usepackage{floatrow}
\newfloatcommand{capbtabbox}{table}[][\FBwidth]
\usepackage{blindtext}
\usepackage{hyperref}
\hypersetup{
    colorlinks   = true, 
    urlcolor     = blue, 
    linkcolor    = red, 
    citecolor    = green
}
\floatstyle{plaintop}
\restylefloat{table}

    \title{\LARGE \bf Variable-state Latent Conditional Random Fields \\
for Facial Expression Recognition and Action Unit Detection}
\author
{
    \parbox{16cm}
    {
        \centering{\large Robert Walecki$^1$, Ognjen Rudovic$^1$, Vladimir Pavlovic$^2$ and Maja Pantic$^1$}\\
        {
            \normalsize$^1$ Computing Department, Imperial College London, UK\\
            $^2$ Department of Computer Science, Rutgers University, USA 
        }
    }
}

\begin{document}

\begin{abstract}
Automated recognition of facial expressions of emotions, and detection of facial action units (AUs), from videos depends critically on modeling of their dynamics. These dynamics are characterized by changes in temporal phases (onset-apex-offset) and intensity of emotion expressions and AUs, the appearance of which may vary considerably among target subjects, making the recognition/detection task very challenging. The state-of-the-art Latent Conditional Random Fields (L-CRF) framework allows one to efficiently encode these dynamics through the latent states accounting for the temporal consistency in emotion expression and ordinal relationships between its intensity levels, these latent states are typically assumed to be either unordered (nominal) or fully ordered (ordinal). Yet, such an approach is often too restrictive. For instance, in the case of AU detection, the goal is to discriminate  between the segments of an image sequence in which this AU is active or inactive. While the sequence segments containing activation of the target AU may better be described using ordinal latent states (corresponding to the AU intensity levels), the inactive segments (i.e., where this AU does not occur) may better be described using unordered (nominal) latent states, as no assumption can be made about their underlying structure (since they can contain either neutral faces or activations of non-target AUs). To address this, we propose the variable-state L-CRF (VSL-CRF) model that automatically selects the optimal latent states for the target image sequence, based on the input data and underlying dynamics of the sequence. To reduce the model overfitting either the nominal or ordinal latent states, we propose a novel graph-Laplacian regularization of the latent states. We evaluate the VSL-CRF on the tasks of facial expression recognition using the CK+ dataset, and AU detection using the GEMEP-FERA and DISFA datasets, and show that the proposed model achieves better generalization performance compared to traditional L-CRFs and other related state-of-the-art models. 
\end{abstract}

\maketitle

\section{Introduction}
\label{sec:introduction}
Facial behavior is believed to be the most important source of information when it comes to affect, attitude, intentions, and social signals interpretation. Machine understanding
of facial expressions could revolutionize user interfaces for artifacts such as robots, mobile devices, cars, and conversational agents. Other valuable applications are in the
domain of medicine and psychology, where it can be used to improve medical assistance as well as develop automated tools for behavioral research \cite{RudovicEtAlPAMI14}. Therefore, automated analysis of facial expressions has attracted a significant research attention \cite{pantic2000automatic}. Facial expressions (FE) are typically described at two levels: the facial affect (emotion) and facial muscle actions (AUs), which stem directly from the message and sign judgment approaches for facial expression measurement \cite{CohnEkman05}. The message judgment approach aims to directly decode the meaning conveyed by a facial display (e.g., in terms of the six basic emotions). Instead, the sign judgment approach aims to study the physical signal used to transmit the message (such as raised cheeks or depressed lips). To this end, the \emph{Facial Action Coding System} (FACS)~\cite{Ekman2002} is used as a gold standard. It is the most comprehensive, anatomically-based system for encoding facial expressions by describing the facial activity based on the activations of 33 AUs. These AUs, individually or in combinations, can describe nearly all-possible facial movements \cite{chu2013selective}.
\begin{figure}[ht]
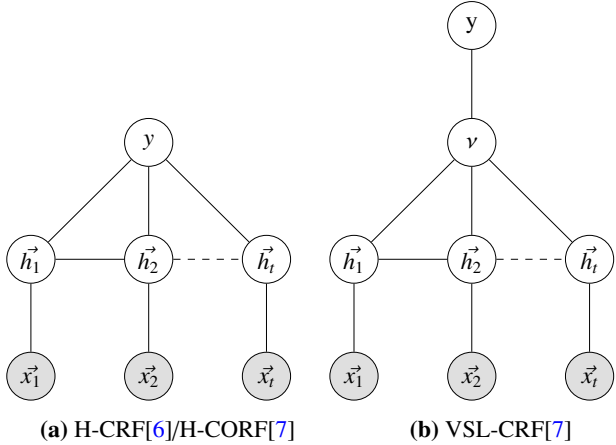

    \centering
    \begin{subfigure}[b]{.48\linewidth}
        \resizebox{0.9\textwidth}{!}{%
            \centering
            \tikz{
                \node[obs] (x1) {$\vec{x_1}$}; %
                \node[obs, right=of x1] (x2) {$\vec{x_2}$}; %
                \node[obs, right=of x2] (xt) {$\vec{x_t}$}; %
                \node[latent, above=of x1] (h1) {$\vec{h_1}$}; %
                \node[latent, above=of x2] (h2) {$\vec{h_2}$}; %
                \node[latent, above=of xt] (ht) {$\vec{h_t}$}; %
                \node[latent, above=of h2, xshift=0cm] (y) {$y$} ; %
                \edge[-] {x1} {h1} ; %
                \edge[-] {x2} {h2} ; %
                \edge[-] {xt} {ht} ; %
                \edge[-] {h1} {h2} ; %
                \edge[-,dashed] {h2} {ht} ; %
                \edge[-] {h1} {y} ; %
                \edge[-] {h2} {y} ; %
                \edge[-] {ht} {y} ; %
            }
        }
        \caption{H-CRF\cite{hcrf06}/H-CORF\cite{hcorf}}
        \label{fig:H-CRF}
    \end{subfigure}%
    \begin{subfigure}[b]{.48\linewidth}
        \resizebox{0.9\textwidth}{!}{%
            \centering

            \tikz{
                \node[obs] (x1) {$\vec{x_1}$}; %
                \node[obs, right=of x1] (x2) {$\vec{x_2}$}; %
                \node[obs, right=of x2] (xt) {$\vec{x_t}$}; %
                \node[latent, above=of x1] (h1) {$\vec{h_1}$}; %
                \node[latent, above=of x2] (h2) {$\vec{h_2}$}; %
                \node[latent, above=of xt] (ht) {$\vec{h_t}$}; %
                \node[latent, above=of h2, xshift=0cm] (nu) {$\nu$} ; %
                \node[latent, above=of nu, xshift=0cm] (y) {y} ; %
                \edge[-] {x1} {h1} ; %
                \edge[-] {x2} {h2} ; %
                \edge[-] {xt} {ht} ; %
                \edge[-] {h1} {h2} ; %
                \edge[-,dashed] {h2} {ht} ; %
                \edge[-] {h1} {nu} ; %
                \edge[-] {h2} {nu} ; %
                \edge[-] {ht} {nu} ; %
                \edge[-] {nu} {y} ; %
            }

        }
        \caption{VSL-CRF\cite{hcorf}}
        \label{fig:H-CRF}
    \end{subfigure}
    \caption{The graph structure of the (a) traditional Latent CRF models H-CRF/H-CORF, and (b) proposed VSL-CRF model. In H-CRF/H-CORF, the latent states $h$, relating the observation sequence ${\bf x}=\{x_1,\dots,x_T\}$ to the target label $y$ (e.g., emotion or AU activation), are allowed to be either nominal or ordinal, while in VSL-CRF the latent variable $\nu=\{nominal,ordinal\}$ performs automatic selection of the optimal latent states for each sequence.}
    \label{fig:graph}
\end{figure}

Early research on facial expression analysis focused mainly on recognition of prototypic facial expressions of six basic emotions (anger, happiness, fear, surprise, sadness, and disgust) and detection of AUs from static facial images \cite{pantic2000automatic}. However, recognizing facial expressions from videos (i.e., image sequences) is more natural and has proved to be more effective~\cite{liu2013learning}. These is motivated by the fact that facial expressions can better be described as a dynamic process that evolves over time. For instance, facial expressions of emotions and AUs undergo a transition of their temporal phases (onset-apex-offset) during the expression development. Similarly, the activation of AUs spans different time intervals that reflect variation in their intensity, as described by FACS. Several works in the field (e.g., \cite{pantic2000automatic,RudovicEtAlPAMI14}) have emphasized the importance of modeling these dynamics for increasing the recognition performance in the target tasks compared to the static methods (see also \cite{zeng2009survey}). 

Most of the state-of-the-art approaches for modeling facial expression dynamics are based on variants of Dynamic Bayesian Networks (DBN) (e.g., Hidden Markov Models (HMM) \cite{Bishop2006Pattern} and Conditional Random Fields (CRF) \cite{Lafferty2001Conditional}). These methods are detailed in Sec. \ref{fer}. In what follows we focus on hierarchical extensions of CRFs \cite{hcorf, Rudovic2012cvpr,RudovicEtAlPAMI14, liu2013learning}, as they are directly related to the model proposed in this paper. These methods can be cast as variants of the CRFs called Latent CRFs (L-CRF) \cite{hcrf06}, and they have also been successfully used in other vision-domains  (e.g. gesture recognition \cite{hcrf06} and human motion estimation \cite{sminchisescu2006conditional}) to encode dynamics of the target tasks. In the context of facial expressions, L-CRFs have been used to model temporal dynamics of facial expressions as a sequence of latent states, relating the image features to the class label (e.g., an emotion category). A typical representative of these models is the Hidden CRF (H-CRF) \cite{quattoni2007hidden, hcrf06,Jain2011hcrf,chang2009learning}, used for facial expression recognition of six basic emotions. Apart from temporal constraints imposed on its latent states, this model fails to account for the ordinal relationships between the latent states. However, the latter may turn important for the model's performance when prior knowledge about the task is available (as in the case of facial expression activations, the intensity of which changes over time). To this end, the recently proposed Hidden Conditional Ordinal Random Field (H-CORF) model \cite{hcorf, Rudovic2012cvpr} imposes additional constraints on the latent states of emotions by exploiting their ordinal relationships. Specifically, this model implicitly enforces the latent states (of emotions) to correlate with temporal phases (or intensity) of emotions by representing them on an ordinal scale. This, in turn, results in the model with fewer parameters, which is less prone to overfitting, and, thus, able to discriminate better between facial expressions of different emotions~\cite{hcorf, Rudovic2012cvpr}.

However, in the L-CRF models such as H-CRF and H-CORF, and their variants, the latent states are assumed to be either nominal or ordinal for each and every class. This representation can be too restrictive since for some classes modeling the latent states as ordinal may help to better capture the structure of the states, i.e., their ordinal relationships, allowing the model to better fit the data. By contrast, it would be wrong to impose ordinal constraints on latent states of the classes that do not exhibit ordinal structure. In this case, the more flexible nominal model will better fit the data. For example, in recognition of emotion-specific expressions, and/or detection of target AU,  we expect the latent states used to model the activation of facial expressions of target emotion class (e.g., happiness) to be correlated with its temporal phases defined on an ordinal scale (neutral$<$onset$<$apex). Similarly, for an AU activation, the latent states should be correlated with its intensity levels, as defined on the Likert scale using FACS (i.e., neutral$<$A$<$B$<$C$<$D$<$E). On the other hand, image sequences of the negative class, i.e., containing a neutral face (without facial activity) or a mix of other non-target facial expressions (different emotions or AUs), are expected to be better fit using nominal states. This is due to the lack of the ordinal structure as well as high variability (activations of various non-target AUs) in such data. We can even go a step further by assuming that the nature of the latent states depends not only on the type of the emotion/AU class (active vs inactive), but that it can also vary for each image sequence of the target classes. For instance, this can occur due to differences in facial expressiveness of different subjects, resulting in the clustering effects of the features caused by the subject-specific variation dominating that related to the facial expressions, and noisy image features (due to the tracking errors in the case of facial landmarks). In these cases, the ordinal relationships could be altered, and, thus, modeling of the ordinal latent states may not be flexible enough to account for the increased levels of variation in the data. To mitigate this, the model should automatically infer what type of the latent states should be used for modeling the dynamics of the input/output data. To this end, we generalize the L-CRF models by relaxing their assumption that the latent states within the target sequence need only be nominal or ordinal. We do so by allowing the model to use both types of latent states for modeling sequences within and across the target classes. Specifically, we introduce a novel latent variable within the L-CRF framework, the state of which defines what type of latent states are best suited for target image sequences. The learning in the proposed model is performed using two newly defined approaches based on max-polling of the latent states, as well as an Expectation-Maximization (EM) algorithm. To reduce potential redundancy in the modeling of the underlying dynamics of facial expressions, we propose the graph-Laplacian regularization of the model parameters that is defined directly on posterior distributions of the latent states. 

The contributions of the proposed work can be summarized as follows:
\begin{itemize}
  \item[1)] We introduce a novel Variable-state L-CRF (VSL-CRF) model for classification of image sequences that, in contrast to existing L-CRF models, has flexibility to use either nominal or ordinal latent states for modeling the underlying dynamics of target sequences. Also, the proposed model selects automatically the optimal latent states for each target sequence.
  \item[2)] We propose two novel learning algorithms based on max-pooling and the EM-like learning of the latent states, as well as graph-Laplacian regularization of the model parameters, for efficient training of the proposed VSL-CRF model. This results in a model that is less prone to overfitting of target data compared to when traditional maximum-likelihood learning (ML) approach is used, as in L-CRF models such as H-CRF and H-CORF. 
  \item[3)] We show on three publicly available datasets (CK+, GEMEP-FERA and DISFA) that by allowing the VSL-CRF model to automatically select the optimal latent states, it can better  learn the underlying dynamics of target facial expressions. This, in turn, results in its superior performance in the sequence-based classification of facial expressions of six basic emotions and detection of target AUs, as well as similar or better frame-based detection of AUs, compared to existing L-CRF models and related state-of-the-art models for the target tasks.  
\end{itemize}
The rest of the paper is organized as follows. Sec. \ref{sec:rel} describes the recent advances in the sequence- and frame-based classification of facial expressions of emotions and AU detection. Sec. \ref{sec:meth}
introduces the proposed methodology. Sec. \ref{sec:exp} describes the conducted experiments and presents the evaluation results, and Sec. \ref{sec:conc} concludes the paper. 

\section{Related Work}


\label{sec:rel}
\subsection{Facial Expression Recognition}
\label{fer}
Facial expression recognition methods can be categorized into the static and dynamic approaches (see \cite{zeng2009survey} for a detailed overview).
The static approach attempts the expression recognition from a single image (typically, the apex of the expression) \cite{shan2005conditional, bartlett2003real, pantic2004facial}. For example, \cite{zhong2014learning} proposed a two-stage multi-task sparse learning framework  to efficiently locate the most discriminative facial patches for the expression classification. The SVM classifier is then used to classify the patches into the six basic emotion categories. The approach in \cite{EF} exploits ensemble of features comprising of
Hierarchical Gaussianization (HG), Scale Invariant Feature Transform (SIFT) and Optic Flow, followed by the SVM-based classification of emotion expressions.

However, a natural facial event such as facial expression of an emotion is dynamic, i.e., it evolves over time by (typically) starting from a neutral expression, followed by its onset, apex, and then the offset, followed by the neutral expression again. For this reason, facial expression recognition from videos is more common than from static images. Although some of the static methods use the features extracted from a window around the target frame, in order to encode dynamics of facial expressions, models for dynamic classification provide a more principled way of doing so. As we mentioned in Sec.\ref{sec:introduction}, most of the dynamic approaches to classification of facial expressions are based on variants of DBNs  such as HMMs and CRFs. For example, \cite{Shang2009nphmm} trained independent HMMs for each emotion category, and then performed emotion classification by comparing the likelihoods of the emotion-specific HMMs. However, discriminative models based on CRFs \cite{Jain2011hcrf,chang2009learning,sebe2007authentic} have been shown to be more effective for the facial expression classification. Furthermore, \cite{wang2013capturing} have shown that capturing more complex time-dependences in the data (beyond the first order dependences as done in linear-chain CRFs) can enhance the facial expression classification performance. Similarly, \cite{Jain2011hcrf,sebe2007authentic} used a generalization of the linear-chain CRF model, a Hidden Conditional Random Field (H-CRF) \cite{hcrf06}, with additional layer of (hidden) variables used to model temporal dynamics of facial expressions. The training of the model was performed using image sequences, but classification of the expressions was done by selecting the most likely class (i.e., emotion category) at each time instance. The authors showed that: (i) having the additional layer of hidden variables results in the model being more discriminative than the standard linear-chain CRF, and (ii) that modeling of the temporal unfolding of the facial shapes is more important than their spatial variation for discriminating between facial expressions of different emotion categories (based on comparisons with SVMs). Another modification of H-CRF, named partially-observed H-CRF, was proposed in \cite{chang2009learning}, where additional hidden variables are added to the model to encode the occurrence of subsets of AU combinations in each image frame, and which are assumed to be known during learning. This method outperformed the standard H-CRF, which does not use a prior information about the AU co-occurrences. In contrast to these models, which still perform per-frame classification of target expressions, \cite{hcorf,Rudovic2012cvpr} proposed the Hidden Conditional Ordinal Random Field (H-CORF) models for the sequence-based classification of facial expressions of emotions and their temporal phases (onset-apex-offset) simultaneously. These models encode ordinal relationships between the temporal phases of emotion expression using either supervised or unsupervised learning of the latent states (corresponding to the temporal phases). The authors showed that improved facial expression recognition can be achieved due to the ordinal modeling of the latent states, with the supervised modeling of the latent states (i.e., using the labels for the temporal phases of emotion expression) outperforming the unsupervised modeling, as expected in this task. Nevertheless, the main limitation of the models listed here is that they restrict their latent states to be either nominal (H-CRF) or ordinal (H-CORF), which may be suboptimal in some cases, as discussed in Sec. \ref{sec:introduction}.
\ADD{
}

\subsection{Facial AU Detection}
As for facial expression recognition, two main approaches are typically adopted for AU detection: static and dynamic modeling.
In the former, image features are extracted from each frame and then fed into a static classifiers such as SVM or AdaBoost \cite{ValstarEtAl05} specifically designed for detection of each AU independently. A more advanced static AU detector, named The Selective Transfer Machine (STM)  \cite{chu2013selective}, has shown great improvements over standard SVMs in the target task. It personalizes the generic SVM classifier by learning the classifier and re-weighting the training samples that are most relevant to the test subject during inference. However, a limitation of this approach is that the re-learning of the target AU detectors has to be performed for each test subject. The modified correlation filter (MCF) \cite{chew2012improved} is also an approach similar in spirit to SVMs and correlation filters, but with the key difference of optimizing only a single hyperplane. This results in more robust AU detection compared to standard SVMs when sequence-level AU labels are used for the frame-based AU detection. The authors of \cite{MKL} proposed a multi-kernel-learning (MKL) approach to AU detection, where they investigate the fusion of different appearance-based image features via the sum of histogram-based kernel functions. These kernels are then used in the SVMs trained for each AU.  To include the temporal information, the authors extract features within AU-specific windows around the image frames used for detection of target AUs. \cite{zhu2014multiple} proposed a multi-task feature learning (MTFL) method for joint AU detection. The MTFL approach and Bayesian networks are used to model AU dependences at both feature and label level, and, thus, perform sjoint AU detection in a probabilistic fashion. Likewise, \cite{zhang2014p} introduces the lp-norm regularization to the MKL, in order to fuse multiple features (using various kernels) and account for the AU-dependencies. Bayesian graphical models were also used to encode sparsity and statistical co-occurrence of AUs \cite{songExploiting} for their joint modeling.

While the methods listed above focus on finding the most discriminative feature representations and/or on inference methods for joint AU detection, they fail to account for temporal information, i.e., AU dynamics. Methods that do so attempt using either temporal image features \cite{koelstra2010dynamic,jiang2012facial} or DBN-based models such as HMMs \cite{valstar2012fully} and CRFs \cite{vanderMaaten12}. In general, these works perform either majority voting using the static detection \cite{ValstarEtAl05}, or detection of the temporal phases of AUs followed by the rule-based classification of the sequences (by detecting the onset-apex-offset sequence of an AU) \cite{valstar2012fully,jiang2014dynamic}. Other temporal models are based on Ordinal CRFs have been proposed for modeling of AU temporal phases \cite{Rudovic2012eccv}, and their intensity \cite{RudovicEtAlPAMI14}, however, they do not perform AU detection. Another approach, termed Cascade of Tasks (CoT) \cite{ding2013facial}, is trained on sequences and applies segment-based detection of AUs. This approach is a combination of three algorithms for static-frame-level-detection, segment-level-detection and transition-level detection. The Interval Temporal Bayesian Networks \cite{wang2013capturing} (ITBN) have also been proposed to capture complex temporal relations among facial events, and for AU detection. The network also represents the spatial dependences among the facial events with a larger variety of time-constrained relations.

Note that the above-mentioned approaches for facial expression recognition and AU detection use either static/dynamic classifiers which are designed for either nominal or ordinal data. While the former imposes no spatial constraints on target classes, the latter does so for all classes (e.g., all emotions are modeled by imposing ordinal constraints).
In the context of the temporal models based on CRFs, this results in the models that are either under-constrained (e.g., H-CRF\cite{hcrf06}) or over-constrained (H-CORF\cite{hcorf}), which limits their representational power. In relation to the state-of-the-art methods, the proposed VSL-CRF model focuses on two key aspects of the facial expression recogniton/ AU detection: (i) modeling of their temporal dynamics (via novel latent states of the L-CRF models) to improve the recognition/detection performance of existing graph-based dynamic models for the target task. (ii) The application of the model to the sequence-based classification and frame-based detection of facial expressions of emotions and AUs. In the following, we introduce the proposed methodology. 

\section{Methodology}
\label{sec:meth}
\ADD{
In this section, we first give a short introduction to ordinal and nominal CRFs, and their L-CRF extensions. We  then introduce the VSL-CRF method that generalizes these approaches. Lastly, we introduce different methods for the model optimization, including the posterior regularization of the latent states.
}
\subsection*{Notation} 
We consider a $K$-class classification problem, where we let $y \in \{1,...,K\}$ be the class label (e.g., emotion category). Each class $y$ is further represented with a sequence of (latent) states denoted as consecutive integers ${ h}\in\{1,\dots,C\}$, where $C$ is the number of possible states (e.g., temporal phases such as neutral-onset-apex of emotion). The sequence of the corresponding image features, denoted by ${ x} =\{ {x}_1 \dots {x}_T\} \in T \times D$, serves as input covariates for predicting $y$ and ${ h}=(h_1,\dots,h_T)$. The length of sequences $T$ can vary from instance to instance, while the input feature dimension $D$ is constant. If not said otherwise, we assume a supervised setting where we are given a training set of $N$ data pairs $\mathcal{D} = \{(y^i,{ x}^i)\}_{i=1}^N$, which are i.i.d.~samples from an underlying but unknown distribution. 
\subsection{Conditional Random Fields (CRF)}
\label{crfs}

CRFs~\cite{crf01} are a class of log-linear models that represent the conditional distribution $P({ h}|{ x})$ as the Gibbs form clamped on the observation ${ x}$:
\begin{equation}
P({ h}|{ x},{\boldsymbol\theta}) = 
  \frac{1}{Z({ x};{\boldsymbol\theta})}
  e^{s({ x},{ h};{\boldsymbol\theta})}.
\label{eq:crf_gibbs}
\end{equation}
Here, $Z({ x};{\boldsymbol\theta}) = \sum_{{ h} \in \mathcal{H}} e^{s({ x},{ h};{\boldsymbol\theta})}$ is the normalizing partition function ($\mathcal{H}$ is a set of all possible output configurations), and ${\boldsymbol\theta}$ are the parameters\footnote{For simplicity, we often drop the dependency on ${\boldsymbol\theta}$ in notations.} of the {\em score function} (or the negative energy) $s({ x},{ h};{\boldsymbol\theta})$. Note that in this model, the states ${ h}$ are observed and they represent the frame labels. 

\subsection*{Linear Chain Conditional Random Fields (CRF)}
We further assume the linear-chain graph structure  $G = (V,E)$ in the model, described by the {\em node} ($r \in V$) and {\em edge} ($e=(r,s) \in E$) potentials. We denote the node features by ${\boldsymbol\Psi}^{(V)}_r({ x},h_r)$ and the edge features by ${\boldsymbol\Psi}^{(E)}_e({ x},h_r,h_s)$. By letting ${\boldsymbol\theta} = \{{ v},{ u}\}$ be the parameters of the node and edge potentials, respectively, $s({ x},{ h};{\boldsymbol\theta})$ can then be written as the sum:
\begin{eqnarray}
\sum_{r \in V} { v}^{\top} {\boldsymbol\Psi}^{(V)}_r({ x},h_r) + \sum_{e=(r,s) \in E} { u}^{\top} {\boldsymbol\Psi}^{(E)}_e({ x},h_r,h_s).
\label{eq:score_node_edge}
\end{eqnarray}
Although the representation in (\ref{eq:score_node_edge}) is so general that it can subsume nearly arbitrary forms of features, the node/edge features are often defined depending on target task. We limit our consideration to two commonly used types of the node features (nominal/ordinal), which can be represented using a general probabilistic model for static modeling of nominal/ordinal classes. This is achieved by setting the potential at node $r$ as ${ v}^{\top} {\boldsymbol\Psi}^{(V)}_r({ x},h_r) \longrightarrow {\boldsymbol\Gamma}^{(V)}_r({ x},h_r)$, where 
\begin{equation}
\Gamma _r^{(V)}(x,h_r) = \sum\limits_{c = 1}^C {{\rm{I}}(h_r  = c)\cdot log P(h_r  = c|{f }({x}))}.
\label{corf:node}
\end{equation}

The {\it \bf nominal} node potential is then obtained by using the multinomial logistic regression (MLR) model \cite{hcrf06}: 
\begin{equation}
\label{ll:nom}
P(h_r^n = c|{f^n}(x,c)){\mkern 1mu}  = {\rm{ }}\frac{{\exp ({f^n}(x,c))}}{{\sum\nolimits_{l = 1}^C {\exp ({f^n}(x,l))} }}{\mkern 1mu} {\mkern 1mu} {\mkern 1mu},
\end{equation}
where $f_n(x,c) = \beta_c^T\cdot[1,x]$, for $c=1,...,C$, and $\beta_c$ is the separating hyperplane for the $c$-th {\it nominal} state of the target class. By plugging the likelihood function in (\ref{ll:nom}) into the node potential in (\ref{corf:node}), we obtain the node features of the standard CRF model.

Recently, several authors proposed using the ranking likelihood to define the {\it \bf ordinal} node potentials. This likelihood is derived from the threshold model for (static) ordinal regression \cite{McCullagh}, and has the form:
\begin{equation}
\label{ll:ord}
P(h_r^{o} = c|f^{o}(x,c)) = \Phi (\frac{ {{b_c} - f^{o}({{x}})}}{\sigma }) - \Phi (\frac{ {{b_{c - 1}} - f^{o}({{x}})}}{\sigma }),
\end{equation}
where $\Phi(\cdot)$ is the standard normal cumulative density function (c.d.f.), and ${f^o}(x) = {a^T} x$. The parameter vector $a$ is used to project the input features onto an {\it ordinal} line divided by the model thresholds or cut-off points ${b_0} =  - \infty  \le  \cdots  \le {b_C} = \infty $, with each bin corresponding to one of the {\it ordinal} states $c=1,...,C$ in the model. The ranking likelihood in (\ref{ll:ord}) is constructed by contaminating the ideal model (see \cite{corf} for details) with Gaussian noise with standard deviation $\sigma$. Again, by plugging the likelihood function in (\ref{ll:ord}) into the node potential in (\ref{corf:node}), we obtain the node features of the Ordinal CRF (CORF) model \cite{corf}. 

In both models defined above (the standard CRF and CORF), the edge potentials  ${\boldsymbol\Psi}^{(E)}_e({ x},h_r,h_s)$ are defined in the same way and have the form:
\begin{equation}
\Big[ I(h_r=c \ \wedge \ h_s=l) \Big]_{C \times C} \times \big|{ x}_r-{x}_s\big|,
\label{corf:edge}
\end{equation}
where $I(\cdot)$ is the indicator function that returns 1 (0) if the argument is true (false). The role of the edge potentials is to assure the temporal consistency of the nominal/ordinal states within a sequence.

\subsection*{Latent Conditional Random Fields (L-CRFs)}
\label{lcrf}

While the CRFs introduced in the previous section aim at modeling/decoding of the state-sequence within a single class, the framework of L-CRFs \cite{hcrf04,hcrf06} aims at the sequence level multi-class classification. This is attained by introducing additional node in the graph structure of CRF/CORFs (see Fig.\ref{fig:graph}) representing the class label, where the latent states ${ h}$ are now treated as unknown. Formally, L-CRFs combine the score functions of $K$ CRFs, one for each class $y=\{1,\dots, K\}$, within the following score function:
\begin{equation}
s(y,{ x},{ h};{\boldsymbol\Omega}) = \sum_{k=1}^K I(k=y) \cdot s({ x},{ h};{\theta}_y),
\label{eq:hcrf_score}
\end{equation}
Where $s({ x},{ h};{\theta}_y)$ is the $y$-th CRF score function, defined as in (\ref{eq:score_node_edge}), and ${\boldsymbol\Omega} = \{{\theta}_k\}_{k=1}^K$ denotes the model parameters. With such score function, the joint conditional distribution of the class and state-sequence is defined as:
\begin{equation}
P(y,{{h}}|{{x}}) = \frac{{\exp (s(y,{{x}},{{h}}))}}{{Z({{x}})}}.
\label{eq:hcrf_class}
\end{equation}
The sequence of the states ${ h}=(h_1,\dots, h_T)$ is unknown, and they are integrated out by directly modeling the class conditional distribution:
\begin{equation}
P(y|{ x}) = \sum_{{ h}} P(y,{ h}|{ x}) = \frac{\sum_{{ h}} \exp(s(y,{ x},{ h}))}
  {Z({ x})}.
\label{eq:hcrf_marg}
\end{equation}
 
Evaluation of the class-conditional $P(y|{ x})$ depends on the partition function $Z({{x}}) = \sum\limits_k {{Z_k}} ({{x}}) = \sum\limits_k {\sum\limits_{{h}} {\exp } } (s(k,{{x}},{{h}}))$, and the class-latent joint posteriors $P(k,h_r,h_s|{ x})= P(h_r,h_s|{ x},k) \cdot P(k|{ x})$. Both can be computed from independent consideration of $K$ individual CRFs.  The model with the {\it nominal} node potentials in the score function in (\ref{eq:hcrf_marg}) is termed Hidden CRF (H-CRF) \cite{hcrf06}. Likewise, the model with the {\it ordinal} node potentials is termed Hidden CORF (H-CORF) \cite{hcorf}.

The parameter optimization in the H-CRF/H-CORF models is carried out by maximizing the (regularized) negative log-likelihood of the class conditional distribution in (\ref{eq:hcrf_marg}). Furthermore, to avoid the constrained optimization in H-CORF (due to the order constraints in parameters ${\bf b}$ of the ordinal node potentials), the displacement variables $\gamma_c$, where $b_j = b_1 + \sum_{k=1}^{j-1} \gamma_k^2$ for $j=2,\dots,C-1$ are introduced. So, ${\bf b}$ is replaced by the unconstrained parameters $\{b_1,\gamma_1,\dots,\gamma_{C-2}\}$. Similarly, the positivity of the ordinal scale parameter is ensured by setting $\sigma=\sigma_0^2$. Although both the objectives of H-CRF/H-CORF are non-convex because of the log-partition function (log-sum-exp of nonlinear concave functions), their log-likelihood objective is bounded below by 0 and are both smooth functions. For this, the standard quasi-Newton (such as Limited-memory BFGS) gradient descent algorithms are typically used to estimate the model parameters (we use the former). The model parameters for H-CRF are given by $\theta^{(n)}_y={\beta_1,\dots,\beta_C}$, where $C$ is the number of nominal latent states for class $y=\{1,\dots,K\}$. Likewise, for H-CORF we have  $\theta^{(o)}_y=\{b_1,\gamma_1,\dots,\gamma_{C-2}, \sigma \}$ for each class in $y$.

\subsection{Variable-state Latent Conditional Random Fields (VSL-CRF)}
\label{vslcrf}

In this section, we generalize the H-CRF/H-CORF models by allowing their latent states to be modeled using either nominal or ordinal potentials (latent states) within each sequence. In this way, we allow the model to select in an unsupervised manner the optimal feature functions for representing the target sequences. In what follows, we provide a formal definition of the model, and then explain its learning and inference. 

\subsubsection{VSL-CRF: Model}
\label{vslmodel}

{\bf Definition} (Variable-state Latent CRF) {\it Let {${\bf \nu}=(\nu_1,\dots,\nu_K)$} be a vector of symbolic states or labels encoding the nature of the latent states $h^\nu$ of the $i$-th sequence, $i=1,\dots, N_y$ from class $y=(1,...,K)$, either as nominal ({$\nu_y=0$}) or ordinal ({$\nu_y=1$}). The score function for class $y$ in the VSL-CRF model is then defined as:}
\begin{equation}
s(y,{{x}},{{h}},{ \nu} ;{\bf{\Omega }}) = \left\{ \begin{array}{l}
\sum\limits_{k = 1}^K I (k = y)\cdot s({{x}},{{h}};\theta _y^n),\,\,{\text {if}}\,{\nu _y} = 0\\
\sum\limits_{k = 1}^K I (k = y)\cdot s({{x}},{{h}};\theta _y^o),\,\,{\text {if}}\,{\nu _y} = 1
\end{array} \right.
\label{eq:nodevsl}
\end{equation}
{\it where the nominal ($s({{x}},{{h}};\theta _y^n)$) and ordinal ($s({{x}},{{h}};\theta _y^o)$) score functions represent the sum of the node and edge potentials, as given by (\ref{corf:node}) and (\ref{corf:edge}), respectively. Then, the full conditional probability of the VSL-CRF model is given by:}

\begin{align}
    \label{eq:HVSL_marg}
    P(y|{ x}) &= \sum_{{ h,{\nu} }} P(y,{ h,\nu}|{ x}) = \frac{\sum_{{ h,\nu}} \exp(s(y,{ x},{ h},\nu))}{Z({ x})}\\
    Z({{x}}) &= \sum_{ k, h, \nu}{\exp (s(k,{{x}},{{h}},\nu ))} 
    \label{eq:VSLd}
\end{align}
Note that, in contrast to L-CRF models introduced in Sec.\ref{crfs}, the VSL-CRF performs also integration over the latent variable $\nu$, the state of which (ordinal or nominal) defines the type of the latent states for each sequence of facial expressions. The definition of the VSL-CRF in Eq.\ref{eq:HVSL_marg} allows it to simultaneously fit both ordinal and nominal latent states to each sequence, which may result in the model overfitting. In the following, we introduce two novel learning strategies in order to avoid over-parametrization of the model, i.e., to prevent the model from using redundant nominal and/or ordinal latent states during inference of target sequences. 

\subsubsection{VSL-CRF: Learning and Inference}

\subsubsection*{{\bf Max-pooling of latent states}} The first learning strategy that we propose constraints the latent states to take either nominal or ordinal sequence of latent states per target sequence. This is different from H-CRF/H-CORF where the latent states can be either nominal/ordinal for each and every target class and the sequence. Formally, the conditional probability in Eq.\ref{eq:HVSL_marg} is now given by: 
\begin{align}
    P(y|{{x}}) &= \frac{{\mathop {\max (}\limits_{ \nu} \sum\limits_h {\exp (s(y,{{x}},{{h}},\nu )))} }}{{Z({{x}})}}\\
    Z({{x}}) &= \sum\limits_k {\mathop {\max (}\limits_{ \nu} \sum\limits_h {\exp (s(k,{{x}},{{h}},\nu )))} }
\label{eq:VSL}
\end{align}

The key aspect of this approach is that now the type of the latent states is explicitly constrained to either nominal or ordinal. This, in turn, leads to the following (regularized) loss function of the VSL-CRF model (further in the text, we denote this model as {\bf VSLm}):
\begin{equation}
RLL({\bf{\Omega }}) =  - \sum\limits_{i = 1}^N {\log P({{\bf{y}}_i}|{{\bf{x}}_i};{\bf{\Omega }})}  + {\lambda _{n(o)}}||\theta _{k = 1..K}^{n(o)}|{|^2},
\label{vslcrf:obj}
\end{equation} 
where ${{\bf{\Omega }} = \{ \theta _k^n,\theta _k^o\} _{k = 1}^K}$. We introduce $L$-2 regularization over the parameters of the nominal/ordinal score functions, the effect of which is controlled by ${\lambda _{n}}/{\lambda _{o}}$, which are found using a validation procedure. 

Unfortunately, the objective function of the VSLm model is both non-convex and non-smooth because of the ${\it {max}}$ function in its conditional distribution. Therefore, the gradients of the objective in Eq.\ref{vslcrf:obj} w.r.t. the parameters ${\bf{\Omega }}$ cannot be directly computed. Yet, the nominal/ordinal score functions are both sub-differentiable. We use this property to construct the sub-gradient \cite{held1974validation} of the VSLm objective at ${\bf{\Omega }}$. Essentially, this boils down to computing the following sub-gradients: 
\[ \partial {\bf \Omega}= \nabla \mathop {\max (}\limits_\nu  \sum\limits_h {\exp (s(k,{\bf{x}},{\bf{h}},\nu )))},\,k=1,\dots,K,\]
which are further given by
\[\left\{ \begin{array}{l}
 \partial {\bf \Omega}=\nabla \sum\limits_h {\exp (s({\bf{x}},{\bf{h}},\theta _k^n))} ,\\ \,\,\,\,\,\,\,\,\,\,\,\,\,\,\,\,\,\,\,\,\,\,\,\,\,\,if\,\sum\limits_h {\exp (s({\bf{x}},{\bf{h}},\theta _k^n))}  > \sum\limits_h {\exp (s({\bf{x}},{\bf{h}},\theta _k^o))} \,\\
 \partial {\bf \Omega}=\nabla \sum\limits_h {\exp (s({\bf{x}},{\bf{h}},\theta _k^0))} ,\,\,otherwise.
\end{array} \right.\]
Thus, at a point ${\bf{\Omega }}^*$ where one of the score functions, say nominal, gives a higher score than the ordinal for the given sequence, $\mathop {\max (}\limits_\nu  \sum\limits_h {\exp (s(k,{\bf{x}},{\bf{h}},\nu )))}$ is differentiable and has the gradient $\partial \theta _k^n=\nabla \sum\limits_h {\exp (s({\bf{x}},{\bf{h}};\theta _k^n))}$, while $\partial \theta _k^o=0$. In other words, to find a subgradient of the maximum of the score functions, we choose the score functions that achieves the maximum for the target sequence at the current parameters, and compute the gradient of that score function only. Once this is performed, the gradient derivation is the same as in the H-CRF/H-CORF models (see \cite{hcorf} for more details). 

\begin{table*}
    \footnotesize
    \rowcolors{2}{white}{tablegray}
    \begin{center}
        \begin{tabular}{l|llllll}
            \toprule
            Dataset                           & Subjects & Videos & Frames/Video & Content      & AU annotation              & Expression annotation \\
            \midrule
            CK+\cite{lucey2010extended}        & 123      & 327    & 20           & posed & binary last frame          & per video             \\
            GEMEP-FERA\cite{valstar2011first} & 7        & 87     & 20-50        & acted       & binary per-frame           & -                     \\
            DISFA\cite{DISFA}                 & 27       & 32     & 4845         & spontaneous  & intensity levels per-frame & -                     \\
            \bottomrule
        \end{tabular}
    \end{center}
\end{table*}

\begin{figure*}[ht]
    \centering
    \subcaptionbox{CK+ \cite{lucey2010extended}}{
        \begin{subfigure}{0.315\linewidth}
            \includegraphics[width=0.48\linewidth]{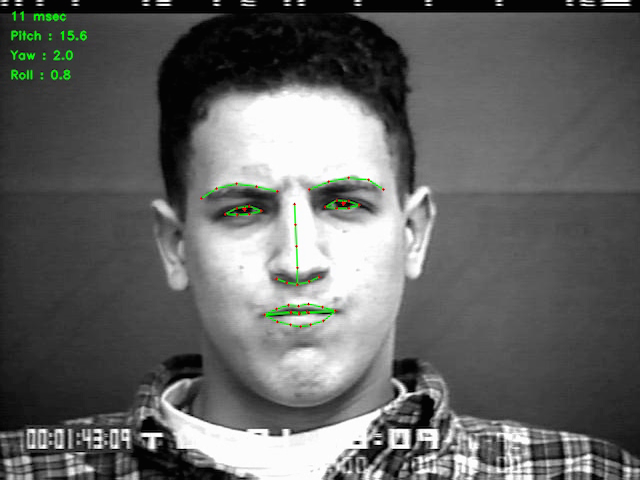}
            \hfill\vspace*{1.2mm}
            \includegraphics[width=0.48\linewidth]{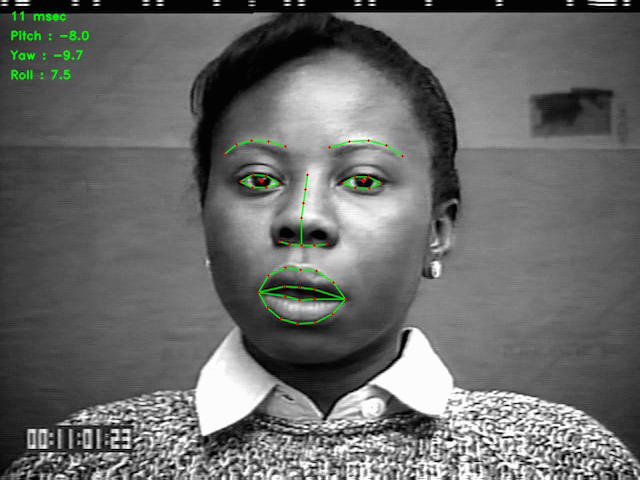}
            \hfill
            \includegraphics[width=0.48\linewidth]{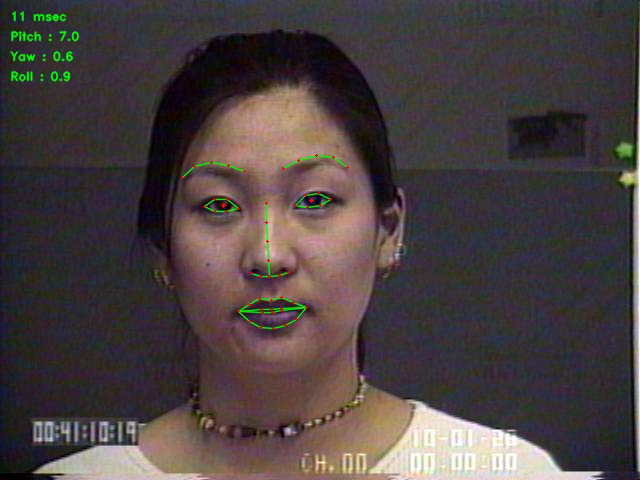}
            \hfill
            \includegraphics[width=0.48\linewidth]{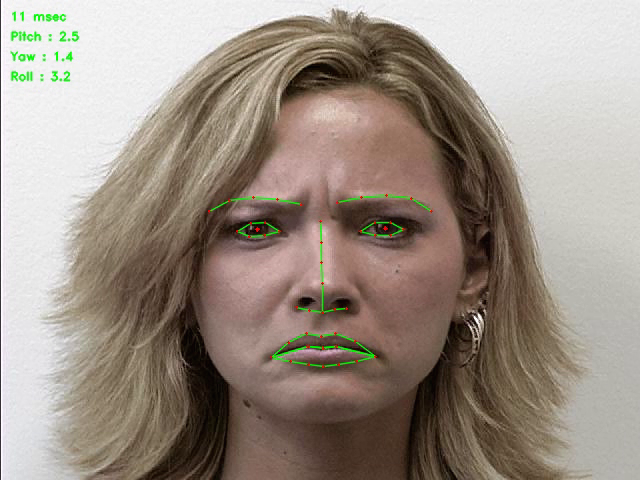}
        \end{subfigure}
    }
    \hfill
    \subcaptionbox{GEMEP-FERA  \cite{valstar2011first}}{
        \begin{subfigure}{0.315\linewidth}
            \includegraphics[width=0.48\linewidth]{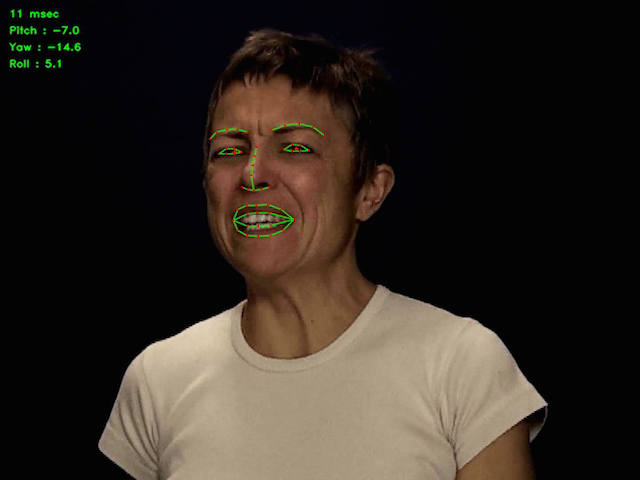}
            \hfill\vspace*{1.2mm}
            \includegraphics[width=0.48\linewidth]{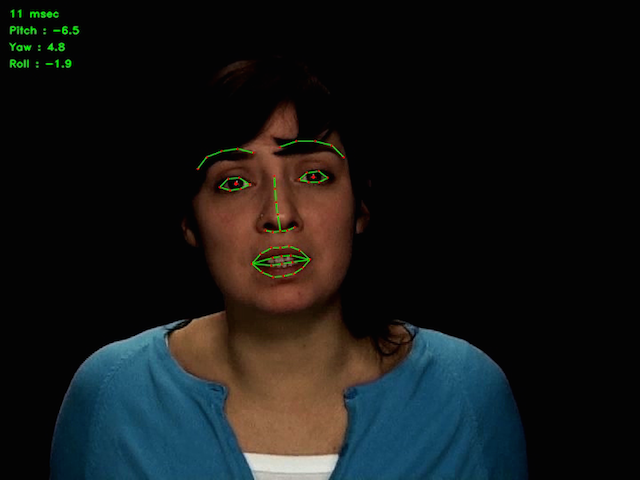}
            \hfill
            \includegraphics[width=0.48\linewidth]{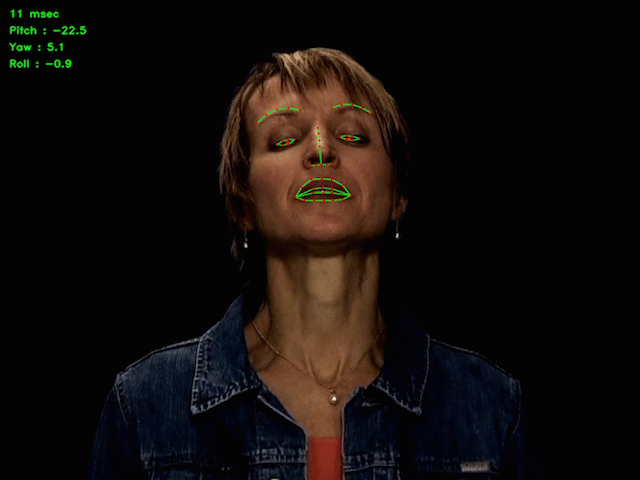}
            \hfill
            \includegraphics[width=0.48\linewidth]{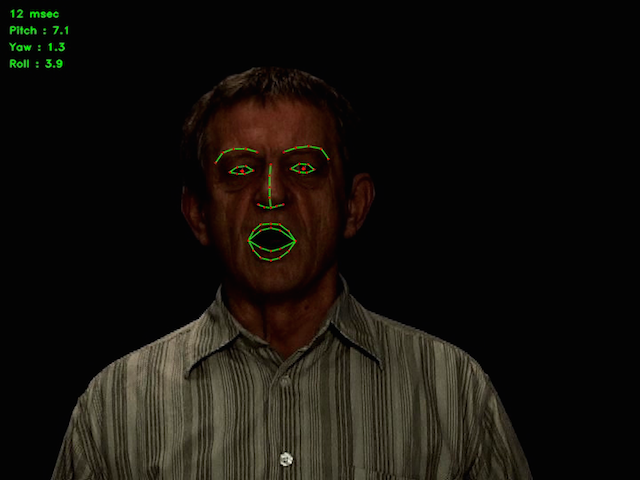}
        \end{subfigure}
    }
    \hfill
    \subcaptionbox{DISFA \cite{DISFA}}{
        \begin{subfigure}{0.315\linewidth}
            \includegraphics[width=0.48\linewidth]{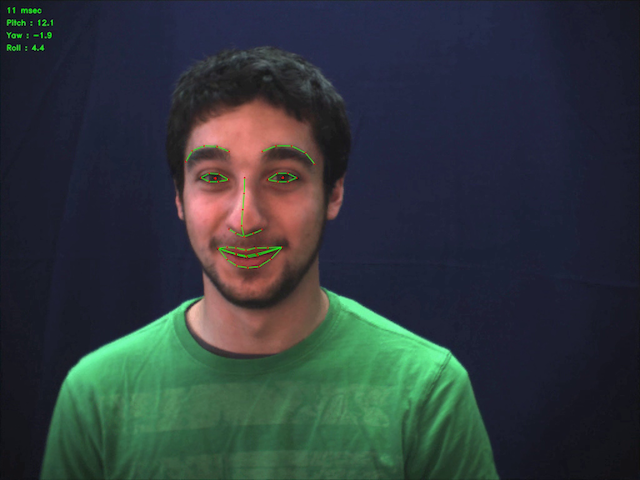}
            \hfill\vspace*{1.2mm}
            \includegraphics[width=0.48\linewidth]{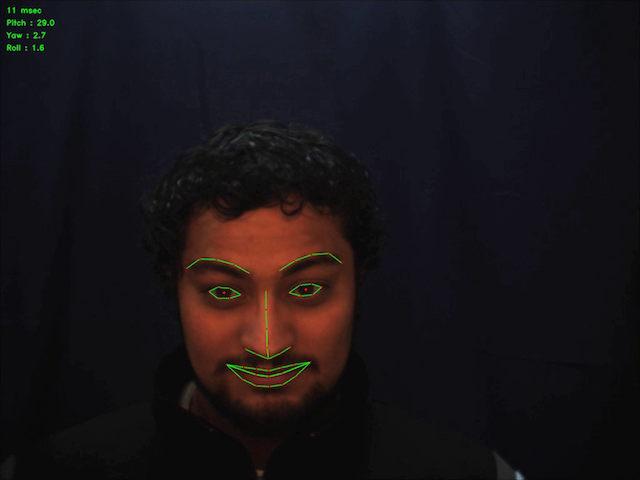}
            \hfill
            \includegraphics[width=0.48\linewidth]{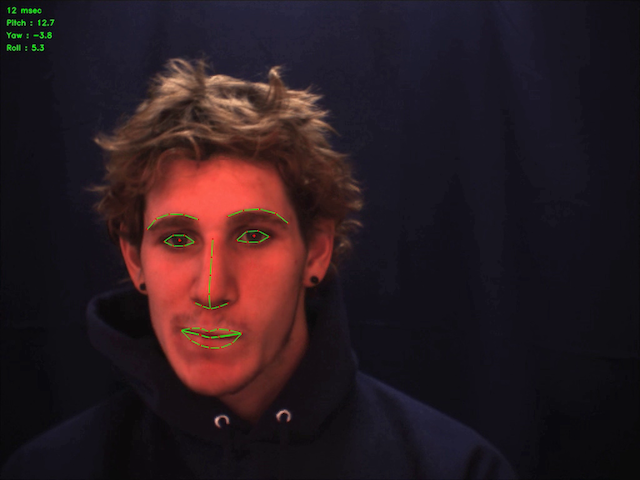}
            \hfill
            \includegraphics[width=0.48\linewidth]{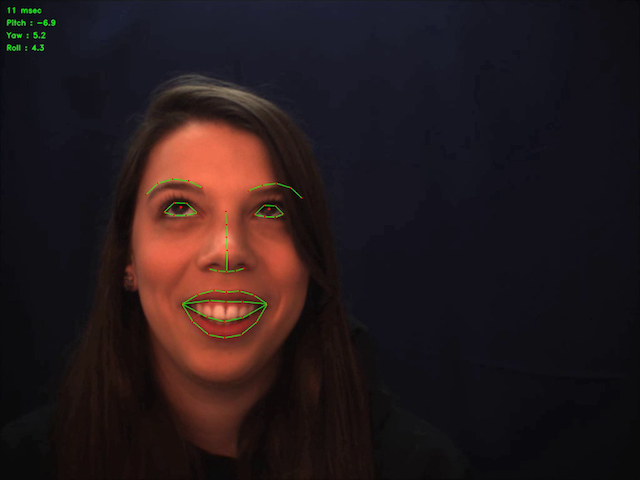}
        \end{subfigure}
    }
    \caption{Sample images with the used facial points from different datasets}
    \label{fig:faces}
\end{figure*}

\subsubsection*{{\bf Fully integrated out latent states}}

The benefits of the VSLm approach are that it prevents the VSL-CRF model from redundant parametrization of the VSL-CRF model that can easily lead to the model overfitting. However, the sub-gradient optimization approach can easily get trapped in the local minimum when searching for the model parameters due to the gradient `switching' caused by the  $max$ function in the objective. To this end, we also employ a learning strategy where both types of the latent states (ordinal and nominal) are fully integrated out, which can be solved using the standard gradient descent optimization as in the existing L-CRF models. Of course, the downside of this is that we may end up with over-parametrization of the target sequences. To remedy this, in addition to direct optimization of the conditional probability, we also introduce an EM approach to the parameter learning. In the proposed EM learning strategy, we exploit the hierarchy in the VSL-CRF model, which allows us to integrate out the latent states $h$ and the indicator variable $\nu$ in an alternating fashion. Note that no empirical studies that investigate the performance of EM vs. the direct optimization in the context of L-CRFs have been reported so far. Furthermore, we introduce novel posterior regularization (see Sec.\ref{posreg}) in the objective function of these approaches, with the aim of implicitly enforcing the model to select either nominal or ordinal latent states for each target sequence during learning\footnote{Note that this regularization does not apply to the VSLm approach as the `hard' selection of the latent states is achieved using the $max$ function.}. Formally, the objective function is given by:
\begin{equation}
    \begin{split}
        RLL({\bf{\Omega }})  =   - \sum\limits_{i = 1}^N {\log P({{\bf{y}}_i}|{{\bf{x}}_i};{\bf{\Omega }})} + {\lambda _{n(o)}}||\theta _{k = 1..K}^{n(o)}|{|^2} + \lambda_{p}\sum_{\nu^{'}}{\mathcal{R}_{\nu^{'} }}
    \end{split}
\label{vslinf}
\end{equation} 
where $P({{\bf{y}}_i}|{{\bf{x}}_i};{\bf{\Omega }})$ is defined by Eq.\ref{eq:VSL}, and $\lambda_{p}$ controls the strength of the posterior regularization defined in Sec.\ref{posreg}. We detail below the two learning approaches.

1. {\it Direct optimization.} Direct optimization of the objective function is performed by minimizing the objective function in Eq.\ref{vslinf} directly w.r.t. all parameters ${\bf \Omega}$ of the model. We denote this approach as {\bf VSLd}. The gradients of the log-likelihood function in the first term on the right side of Eq.\ref{vslinf}  are given by:
\begin{align*}
    \begin{split}
        \frac{\partial log(P(y,\nu|x))}{\partial \Omega}=&
                                                    \mathds{E}_{P(\nu,h|x,y)}[\frac{\partial s(y,\nu,x,h)}{\partial \Omega}]  \\
                                                    -&\mathds{E}_{P(y,\nu,h|x)}[\frac{\partial s(y,x,h)}{\partial \Omega}]
    \end{split}
\end{align*}
The sum of gradient derivations for H-CRF (for $\nu=0$) and H-CORF (for $\nu=1$) can be used to obtain these gradients. The computation of the gradients for the model parameters w.r.t. the regularizers in Eq.\ref{vslinf} is then straightforward.  In all our experiments, we used the Limited-memory BFGS method for the gradient computation.

2. {\it Expectation-Maximization (EM) optimization.}  Alternatively, the model parameter can be obtained using the EM algorithm. The EM algorithm \cite{Bishop2006Pattern} is an iterative optimization approach that can be employed to find the latent state parameters ${\bf {\Omega}}$ that maximize the VSL-CRF objective (Eq.\ref{vslinf}) in two steps. In the E-step, the posterior probability of the binary latent variable $\nu$ is computed as $P(\nu|x,y)$, i.e., by integrating out the latent states $h$, for each target sequence. Then, the maximum-likelihood parameter estimates of the model parameters ${\bf{ \Omega}}$ are computed in the M-step. This process is repeated until the convergence of the objective in Eq.\ref{vslinf}. More specifically, in the E-step, we compute the posterior probabilities for each target sequence using the auxiliary function:

\begin{equation}
q(\nu_i)=p(\nu_i|{\bf y}_i,{\bf x}_i,{\bf \Omega}^j)
\end{equation}
This is followed by the M-step, where a new parameter vector ${\bf \Omega}^{j+1}$ is obtained by maximizing the likelihood function using the current posterior for $\nu$:
\begin{equation}
 \begin{split}
{{\bf{\Omega }}^{j + 1}}= \mathop {\arg \max }\limits_{_{\bf{\Omega }}}  &\sum\limits_{i = 1, \ldots ,N} {\sum\limits_{{\nu _i}} q } ({\nu _i})\log P({{\bf{y}}_i},{\nu _i}|{{\bf{x}}_i},{{\bf{\Omega }}^j})\\
						   & - {\lambda _{n(o)}}||{{\bf{\Omega }}^j}|{|^2} - \lambda_{p}\sum_{\nu^{'}}{\mathcal{R}_{\nu^{'} }}.
\end{split}
\end{equation}
In our experiments, we initialized the model with a uniform distribution $P(\nu=o,k)=0.5$ and $P(\nu=n,k)=0.5$ for all $k$ and ran the  EM-algorithm until it converged. We denote this learning approach as \textbf{VSLem}. It is important to mention that the most important aspect of the \textbf{VSLem} approach, compared to the \textbf{VSLd}, is that the in the latter, the  importance of both nominal and ordinal states is equal and does not change during learning. By contrast, through the E-step, the \textbf{VSLem} dynamically adapts the weight of each model (nominal vs ordinal) for each sequence. Together with the proposed posterior regularization, this is expected to drive the type of latent states for each sequence to either nominal or ordinal, and thus, avoid over-parametrization of the target data. 
\subsection*{{\bf Prediction}} 
\label{pred}
Once the model parameters ${\bf \Omega}$ are learned using either of the proposed approaches (\textbf{VSLm}, \textbf{VSLd} or \textbf{VSLem}), the inference of test data can be performed in two ways, depending on the target task. The first task is sequence-based classification of facial expressions. The goal here is to classify the pre-segmented sequences of facial expressions (e.g., emotions) into one of target classes. In the case of AUs, the goal is to perform detection of the target AU from pre-segmented sequences classified into active (containing activations of the target AU), and `all other' (containing neutral facial expressions and/or facial expressions of non-target AUs). The assignment of a test sequence to the particular class is accomplished by the MAP rule $y^* = \arg \max_{y} P(y|{\bf x}^*)$. In the case of frame-based classification of target facial expressions, the learned models are used to compute the likelihood of each time-window in the input test sequence. Then, the central frame in the window is assigned the target class, as given by the MAP rule mentioned above.
\subsection{Posterior Regularization}
\label{posreg}
In this section, we show how geometric knowledge of the posterior probability distribution can be used in our optimization framework. This is motivated by recent works \cite{ganchev2010posterior,zhu2014bayesian,ganchev2013cross} on posterior regularization in the conditional models, used to improve the parameter learning by incorporating prior knowledge. Formally, let $\Theta$ denote model parameters and $H$ denote hidden variables. Given a set of observed data $\mathcal{D}$, posterior regularization is generally defined as solving a regularized maximum likelihood estimation (MLE) problem:
\begin{align}
P(y|{\bf{x}}) = 
\mathop {\max }\limits_{\Theta} 
    \mathcal{L}(\Theta;\mathcal{D})
    +
\Phi (p(H|{\cal D},\Theta ))
    \label{eg:genPR}
\end{align}
where
$\mathcal{L}(\Theta;\mathcal{D})$
is the marginal likelihood of 
$\mathcal{D}$, and $\Phi(\cdot)$ is a regularization function of the model posteriors over latent variables. A common definition for $\Phi(\cdot)$ is the KL-divergence between a desired distribution with certain properties over latent variables and the model posterior distribution. In this paper, $H$ corresponds to the sequence latent state $\nu$. This parameter is not known and no assumptions can be made in order to construct the KL-divergence. However, we make use of the prior knowledge that sequences, which are sampled from the same class should have the same 
latent states. For instance, we assume that if two sequences $\{{\bf y}^{1},{\bf x}^{1}\}$ and $\{{\bf y}^{2},{\bf x}^{2}\}$ are from the same target class $k$, then the conditional probabilities $P(\nu|{\bf y}^{1}=k,{\bf x}^{1})$ and $P(\nu|{\bf y}^{2}=k,{\bf x}^{2})$ should be similar. Suppose further that there are $K$ classes and let $f_{\nu,k}({\bf x})=P(\nu|{\bf y}=k,{\bf x})$ be the conditional posterior probability density function for each class defined as $P(\nu |{\bf x,y}){\rm{ }} = \sum\limits_h P (h,\nu |{\bf x,y})$. Then, the regularization is performed by minimizing the distance between each element of $f_{\nu}$ having the same class label. This can be solved by using the graph Laplacian $L$ \cite{chung1997spectral} regularization approach. To this end, we construct a graph $G$ in which each node $n_{i}$ corresponds to a sequence ${\bf x}^{i}$ with the class label ${\bf y}^{i}$. We connect all nodes with edges $e_{ij}$ that have the weight $s_{ij}$, which is defined by a similarity matrix $S$.  In this work, we assign value 1, if and only if ${\bf y}^{i}={\bf y}^{j}$, $i,j=1,\dots,N$, and $0$ otherwise. This ensures that only the sequences that come from the same class of facial expressions or contain activation of the same AU, are connected.
Finally, the graph Laplacian is constructed as $L=D-S$, where $D$ is a diagonal matrix, the entries of which are column-sums of $S$, that is, $D_{ij}=\sum_{j}{S_{ij}}$. Then, the proposed posterior regularization $\mathcal{R_{\nu}}$ is defined as follows:
\begin{align*} 
    \mathcal{R_{\nu}}&=\frac{1}{2}\sum_{i,j=1}^{m}{S_{ij}\cdot(P(\nu|y^{i},x^{i})}-P(\nu|y^{j},x^{j}))\\
                     &=\sum_{i=1}^{m}{P(\nu|y^{i},x^{i})^2D_{ij} - \sum_{i,j=1}^{m}{P(\nu|y^{i},{x^{i})P(\nu|y^{j},x^{j})S_{ij} } }}\\
                     &=\vec{f}^{T}_{\nu} D \vec{f}_{\nu}-\vec{f}^{T}_{\nu} S \vec{f}_{\nu}\\
                     &=\vec{f}^{T}_{\nu} L \vec{f}_{\nu}
    \label{eq:LapReg}
\end{align*} 
where
\begin{align} 
\vec{f}_{\nu}=(P(\nu|y^{1},x^{1}),...,P(\nu|y^{m},x^{m}))^T
\end{align}
Note that the larger values of the disparity in  $\vec{f}_{\nu}$ result in a larger regularization loss $\mathcal{R_{\nu}}$ for state $\nu=\{n,o\}$.
The matrix $L$ is positive semi-definite, so $\mathcal{R_{\nu}}$ is convex in $\vec{f}_{\nu}$ and by minimizing $\mathcal{R_{\nu}}$, we get a conditional distribution $f_{\nu}$ which is sufficiently smooth on the data manifold. \\

\section{Experiments}
\label{sec:exp}
\begin{figure*}[ht] 
\footnotesize
\rowcolors{3}{white}{tablegray}
\begin{tabularx}{\linewidth}{l|XXXXXc|XXXXXXXXX}
\rowcolor{tablegray}
\toprule
Emotion  & SVM  & HCRF & HCORF      & VSLm      & VSLd      & VSLem                  & CLM                   & Cov3D                  & ITBN                     & POHCRF                   & STMexp                 & TMS                 & MCSPL                    \\
         & (SB) &      &            &           &           &                       & \cite{chew2011person} & \cite{sanin2013spatio} & \cite{wang2013capturing} & \cite{chang2009learning} & \cite{liu2013learning} & \cite{Jain2011hcrf} & \cite{zhong2014learning} \\
\midrule                                                                                                                                                                                                                                                  
Anger    & 76.7 & 95.5 & 93.3       & 93.3      & \bf{97.8} & 97.8                  & 70.1                  & 94.4                   & 91.1                     & 69.4                     & ---                    & 97.9                & 76.3                     \\
Contempt & 45.3 & 82.4 & 70.6       & 84.2      & 88.2      & 88.2                  & 52.4                  & \bf{100.0}             & 78.6                     & ---                      & ---                    & ---                 & ---                      \\
Disgust  & 82.1 & 94.9 & \bf{98.3}  & \bf{98.3} & 96.6      & 96.6                  & 92.5                  & 95.5                   & 94.0                     & 88.9                     & ---                    & 97.9                & 94.1                     \\
Fear     & 67.4 & 84.0 & 69.2       & \bf{96.0} & 92.0      & \bf{96.0}             & 72.1                  & 90.0                   & 83.3                     & 87.7                     & ---                    & 90.5                & 86.2                     \\
Happy    & 86.2 & 95.6 & 97.1       & 97.1      & 97.1      & 98.6                  & 94.2                  & 96.2                   & 89.8                     & 98.0                     & ---                    & \bf{99.6}           & 96.4                     \\
Sadness  & 62.4 & 64.2 & 79.3       & 87.9      & 87.9      & 87.9                  & 45.9                  & 70.0                   & 76.0                     & \bf{97.5}                & ---                    & 90.1                & 88.3                     \\
Surprise & 87.0 & 98.7 & \bf{100.0} & 98.7      & 97.4      & \bf{100.0}            & 93.6                  & \bf{100.0}             & 91.3                     & 98.6                     & ---                    & 98.9                & 98.7                     \\
\midrule                                                                                                                                                                                                                                                  
Avg      & 72.4 & 87.9 & 86.8       & 93.6      & 93.8      & 95.1 (*\textbf{96.1}) & 74.4                  & 92.3                   & 86.3                     & 90.0                     & 94.2                   & \bf{95.8}           & 90.0                     \\
\bottomrule
\end{tabularx}
\caption{Per-sequence classification rate on the \textbf{CK+} database and comparison with the state-of-the-art.}
\label{tab:ck}
\end{figure*}

In this section, we evaluate performance of the proposed VSL-CRF model and different learning strategies in the tasks of classification of facial expressions of emotions, and AU detection. 
The presented experiments are conducted on three publicly available facial expression datasets: Extended Cohn-Kanade (CK+) \cite{lucey2010extended}, GEMEP-FERA \cite{valstar2011first} and DISFA \cite{DISFA}. We also compare the performance of the proposed models with the state-of-the-art methods for both tasks, in the sequence-based classification and frame-based detection settings.

\subsection{Experimental Setup}
\subsubsection*{Datasets}
The facial expression datasets used in this work are summarized in Table \ref{fig:faces}. The \textbf{CK+} dataset contains 593 facial expression sequences from 123 different subjects. Each sequence begins with a neutral face and ends at the peak intensity of facial expression of target emotion category. In total, 327 sequences that are labeled in terms of the basic emotions: Anger, Contempt, Disgust, Fear, Happiness, Sadness, or Surprise, are used. We performed 10-fold subject-independent cross-validation on this dataset. The \textbf{GEMEP-FERA} dataset contains 87 image sequences of 7 subjects with the per-frame labels for the AU (1,2,4,6,7,10,12,15,17,18,25 and 26) activations (present or not). Furthermore, in the target videos, each participant shows facial expressions of the emotion categories: Anger, Fear, Joy, Relief or Sadness. We report our results using a 7 fold subject-independent cross validation, where each fold contained image sequences of a different subject. The \textbf{DISFA} dataset, contains 32 sequences from 27 subjects. Each sequence in this dataset is 4000 frames long, and each frame is labeled in terms of the intensity level (using FACS) for each AU (1,2,4,5,6,9,12,15,17,20,25 and 26). For our detection approach, we used the frames with the AU intensity higher than 0 as positive examples, and the remaining ones as negative. We performed a 10 fold subject independent cross-validation on this dataset.

\subsubsection*{Sequence-based training}
\label{sec:setup}
The proposed models require sequential data for training and prediction and the CK+ database can be directly used. 
However, the AU databases GEMEP-FERA and DISFA require a pre-segmentation step in order to extract sequence training data from these databases.
We created a training datasets that consists of active and not-active subsequences of each AU. More specifically, from the full dataset, we selected the segments in which the target AU is active (inactive) for the duration of at least 6 frames, and used these as positive (negative) sequences for training. We then balanced the data by removing inactive sequences.
Note that we selected the threshold of 6 frames because less than this consistently downgrades the performance on most target AUs, as can be seen from Fig.\ref{fig:cv_frame}. Once the VSL-CRF models are trained using these pre-segmented data, we apply it in both sequence-based and frame-based manner, as explained in Sec.\ref{pred}.

\subsubsection*{Input Features}
\ADD{
We used the locations of 49 facial points, extracted from target images sequences using the appearance-based facial tracker in \cite{akshay_wild}. Fig. \ref{fig:faces} depicts the used facial points from each dataset as input features. The pre-processing of the features was performed by first applying Procrustes analysis to align the facial points to the mean faces of the datasets. This is important in order to reduce the effects of head-pose and subject-specific variation. We then applied PCA to reduce the feature size, retaining $97\%$ of energy, resulting in 18, 21, and 24 dimensional feature vectors for the CK+, DISFA and GEMEP-FERA datasets, respectively. 
}
\begin{figure*}[t]
\centering
\begin{subfigure}{.35\linewidth}
      \includegraphics[width=\linewidth]{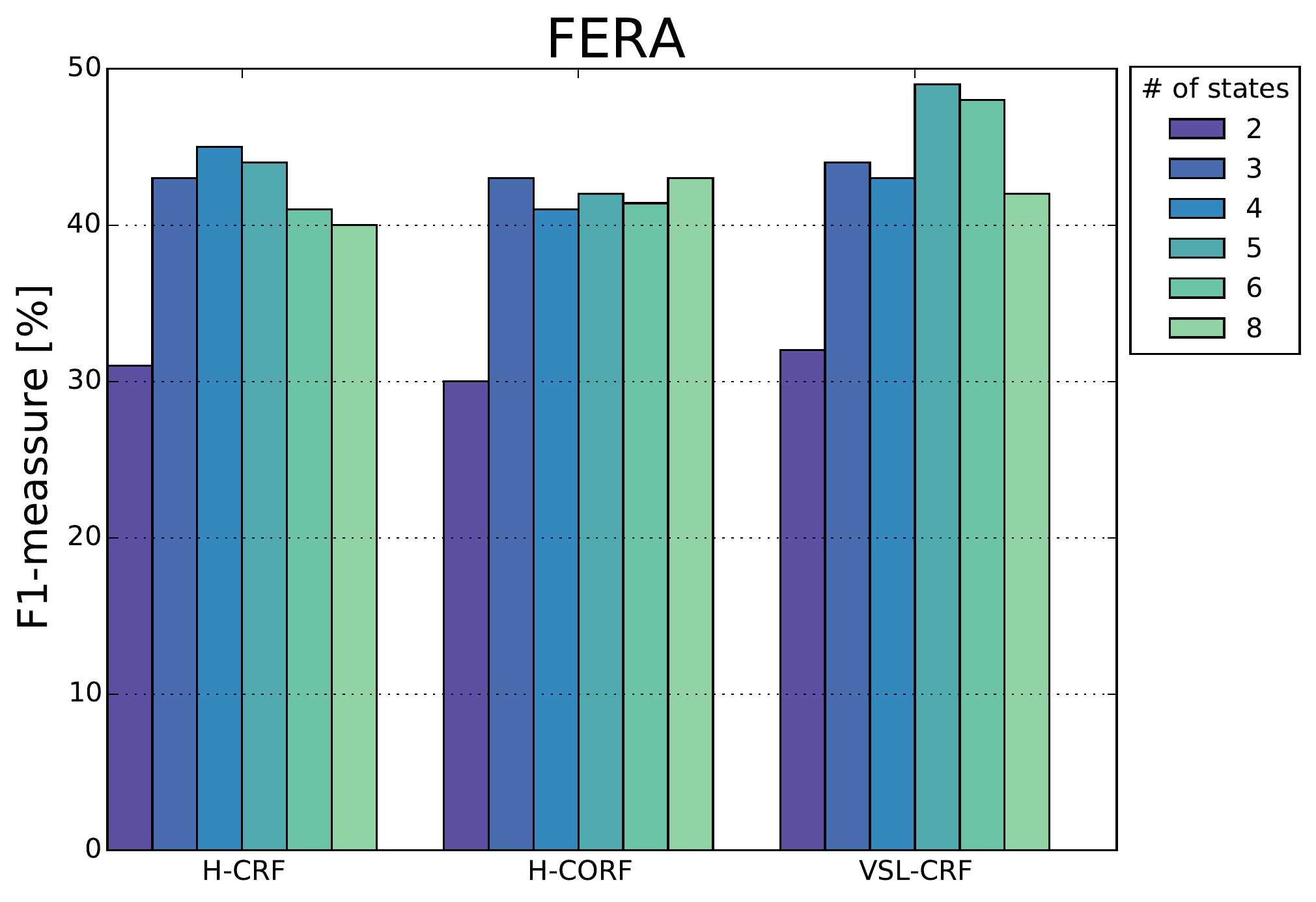}
    \caption{AU6 from the GEMEP-FERA database}
\end{subfigure}%
\hspace{2cm}
\begin{subfigure}{.35\linewidth}
      \includegraphics[width=\linewidth]{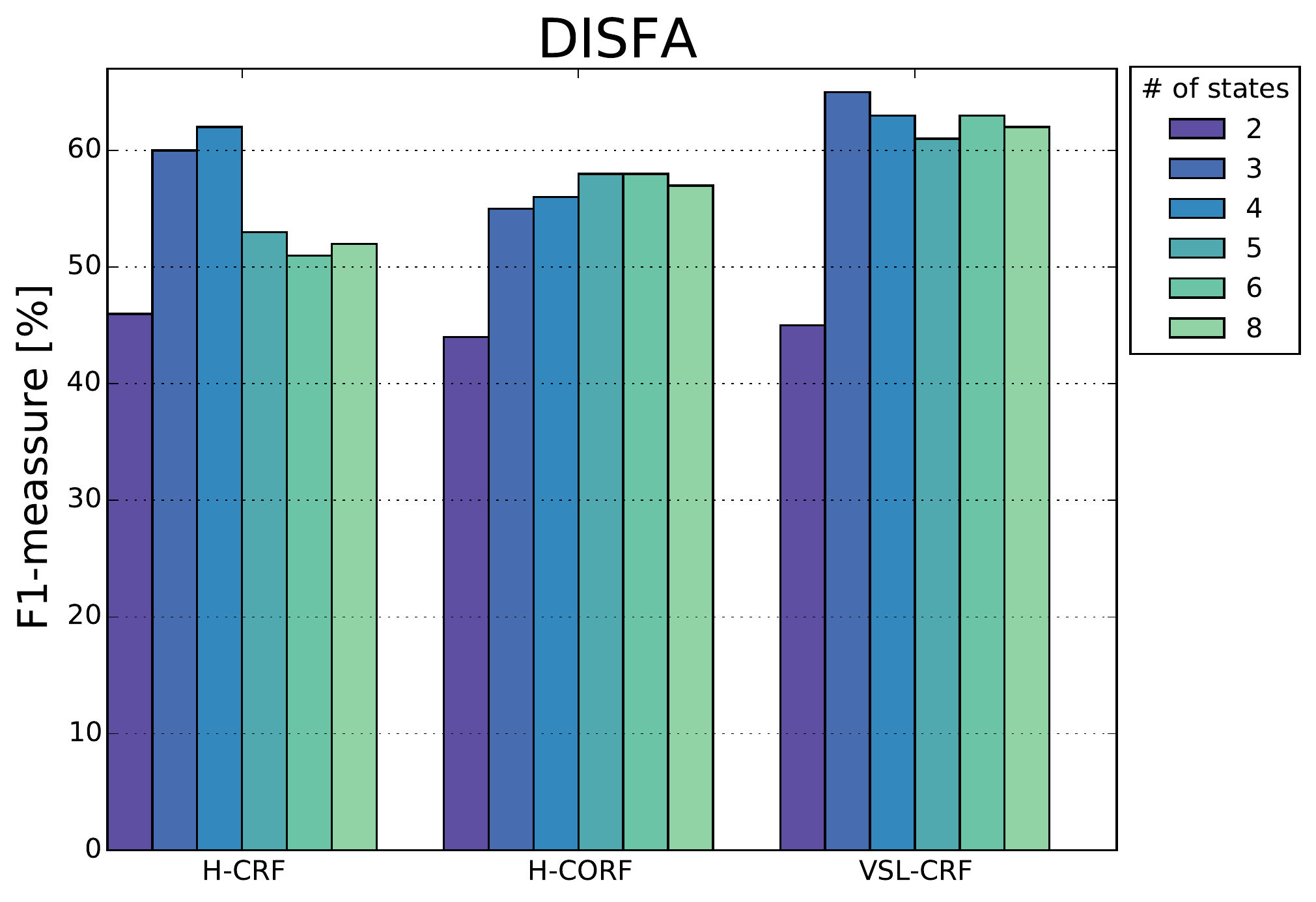}
    \caption{AU6 from the DISFA database}
\end{subfigure}%
\caption{Cross validation over the number of latent states. The tables show the F1-per-sequence measure on AU6 from (a) the GEMEP-FERA and (b) the DISFA datasets w.r.t. the different number of the latent states (nominal and ordinal). In the case of VSL-CRF, the shown number is used separately for nominal and ordinal states.}
\label{fig:cv_states}
\end{figure*}

\subsubsection*{Parameter selection}
The model parameters that need to be pre-defined are the fixed number of latent states $C$ and the regularization parameter  
$\lambda_{o}$,$\lambda_{n}$ and $\lambda_{p}$. We found the optimal number of latent states by applying a grid search over different settings (in a subject-independent manner). In particular, we applied a two fold cross validation on different AUs from target datasets. To illustrate this, in Fig.\ref{fig:cv_states} we show the F1-scores for sequence-based detection of AU6 from the GEMEP-FERA and DISFA datasets when a different number of latent states is used in the compared models: H-CRF, H-CORF and VSLd. The results drop for the H-CRF model when selecting more than 4 latent states per class. This is mainly because of overfitting but also because of the higher dimensionality of the problem. This effect is not significant for the H-CORF model since the ordinal constrains prevent this model from overfitting. However, in all experiments on all AUs, the F1-measure has a strong increase from 2 to 3 hidden states, which is the number of states corresponding to the temporal phases of expression development (neutral-onset/offset-apex). Adding more states does not improve the models' performance significantly but increases their complexity. Therefore, we set in all our experiments the number of hidden states $C=3$ for both ordinal and nominal classes. It is important to mention that although VSL-CRF has more latent states per class (3 nominal and 3 ordinal), as noted above, increasing the number of states in H-CRF and H-CORF does not improve their performance significantly. Consequently, the difference in the performance of the compared models (shown in the experiments below) cannot be attributed to the difference in the number of their latent states. Lastly, the regularization parameters $\lambda_{n/o}$ and $\lambda_{p}$ were set using a grid-search procedure on the validation set found separately for each target fold (no test data were used to perform this validation).

\subsection{Evaluation Measure}
We report the classification/detection results using the standard F1-score. This score is widely used for AU-detection and classification of facial expressions of emotions because of its robustness to the imbalance in positive and negative samples, which is very common in the case of AUs. For each AU, the F1-measure is computed based on a frame-based detection (i.e. an AU detection has to be specified for every frame, for every AU, as being either present or absent). We also provide the results for the sequence-based classification, where the F1-score for sequences is computed based on a sequence-based prediction, and then weighted by the number of frames in each sequence. We do so in order to have the fair comparison with the frame-based approaches. We refer to these metrics F1-sequence-based  for the sequence based approaches, and the F1-frame-based for the frame-based detection. For emotion classification, we used the F1 score, without weighting with the number of frames in the expression sequence, as methods compared on the CK+ dataset perform the sequence-based classification.

\subsubsection{Compared Methods}
In all our experiments, as the baseline for the classification we also include the results obtained by first applying the multi-class SVMs (with the RBF kernel) and trained/evaluated per frame to obtain the F1-frame-based measure. The sequence labels and the F1-sequence-based measure were obtained by majority voting over the frames within the sequence. The results for H-CRF and H-CORF, were obtained using our own implementation\footnote{We provide a toolbox with the Matlab code for the compared H-CRF, H-CORF and VSL-CRF models, at http://ibug.doc.ic.ac.uk/resources/DOC-Toolbox/}. The initial parameters of the models were set using the same approach as in the VSL-CRF. To compare the performance of target models with the state-of-the-art models for each of target tasks (sequence-based emotion recognition and frame-based AU detection), we report the results from the original papers, as detailed below. 


\subsubsection*{Sequence-based Methods}
Note that some of the methods compared use different number of folds when performing cross-validation on the CK+ dataset. Specifically, 
PO-HCRF9 (partially observed H-CRF) \cite{chang2009learning} used a 5-fold cross-validation. In this method, some states are observed during training and represent activations of AUs but the goal is to classify emotions. TMS \cite{Jain2011hcrf} (Temporal Modeling of Shapes) uses Latent-Dynamic CRFs \cite{sebe2007authentic} for a frame-based prediction. However, this predictions are then used to obtain the sequence label. They applied a 4-fold cross validation. ITBN \cite{wang2013capturing} (Interval Temporal Bayesian Network) aims to model temporally overlapping or sequential primitive facial events and the experiments are performed in a 15-fold cross validation setup. Cov3D \cite{sanin2013spatio} is based on spatio-temporal covariance descriptors. The descriptors belong to the group of matrices, which can be formulated as a connected manifold. The authors used a 5-fold cross validation. The Constrained Local Method (CLM) \cite{chew2011person} is a generic or person-independent face alignment algorithm with goal of finding the shape which is described by a 2D trianguleted mesh that fits the target face. They use a 10-fold experimental setup. The MTSL \cite{zhong2014learning} is a multitask sparse learning framework in which expression recognition and face verification tasks, are coupled to learn specific facial patches for individual expression. Lastly, we compare our method to the state-of-the-art method for target task, STM-ExpLet \cite{liu2013learning}. The approach  combines low-level features from videos with a spatio-temporal manifold learning framework and they evaluate the method using 10-fold cross-validation. 

\subsubsection*{Sequence-based Results}

Table \ref{tab:ck} shows the results for facial expression recognition from the CK+ dataset. The average classification rate is obtained by unweighted averaging of the results of the 6 basic emotion (*) plus the contempt emotion. Note that while the results of the compared L-CRF models are directly comparable, as they are trained/tested on the same data/folds, this is not the case with the rest of the models as they use different evaluation settings. However, we report their performance for the sake of comparisons. Note also that in this task, i.e., the classification of facial expressions of emotions, the dynamic methods (H-CRF, H-CORF and VSL-CRF) outperform by the large margin the sequence-based SVM classifier that does not account for temporal dynamics. This table also shows that the proposed variable-state method outperforms the other methods that do not have the flexibility to select the best latent states. On the other hand, the proposed VSLem learning strategy improves the classification performance compared to the other two introduced learning methods (VSLm and VSLd). We attribute this to the iterative learning of the latent states, as well as the posterior regularization, which, evidently, together help to increase the discriminative power of the VSL-CRF model. Lastly, the proposed VSLem achieves the state-of-the-art performance in the target task by performing similar or better than the best performing state-of-the-art models, STM-ExpLet and TMS. 

Tables \ref{tab:disfa_sb} and \ref{tab:fera_sb} show the results for AU detection on the DISFA and GEMEP-FERA database using pre-segmented sequences. Again, the proposed VSL-CRF model outperforms the models that use only nominal (H-CRF) or ordinal (H-CORF) states, trained/tested on identical data/folds. Furthermore, the highest detection rate is again achieved using the VSLem model on both the DISFA and GEMEP-FERA datasets. Moreover, all the VSL-CRF methods achieve significantly higher results than the other L-CRF models, which is mainly because of the ability to select the optimal states per sequence.

\subsubsection*{Frame-based Methods}
\begin{figure*}[t]
\centering
\begin{subfigure}{.35\linewidth}
    \includegraphics[width=\linewidth]{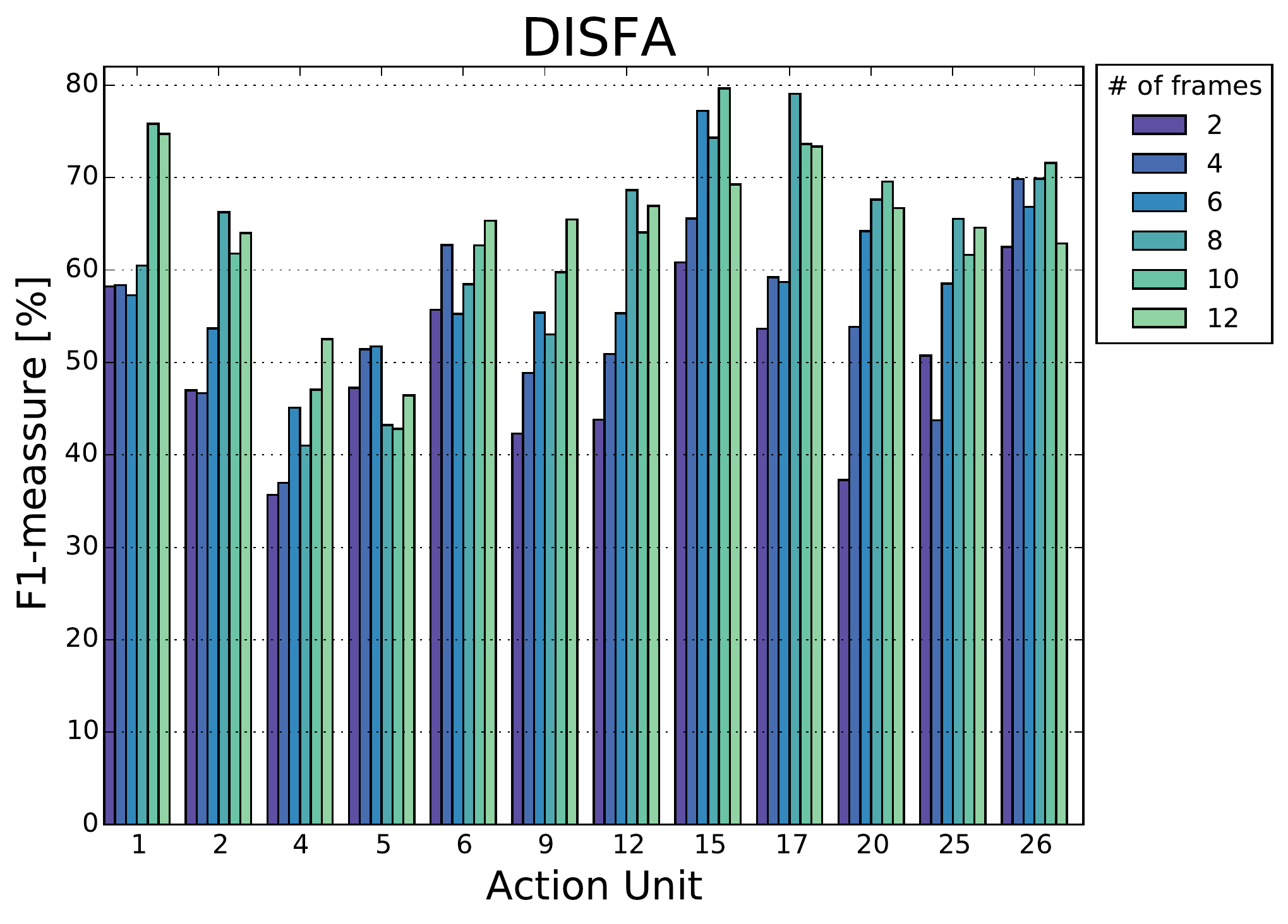}
\end{subfigure}%
\hspace{2cm}
\begin{subfigure}{.35\linewidth}
    \includegraphics[width=\linewidth]{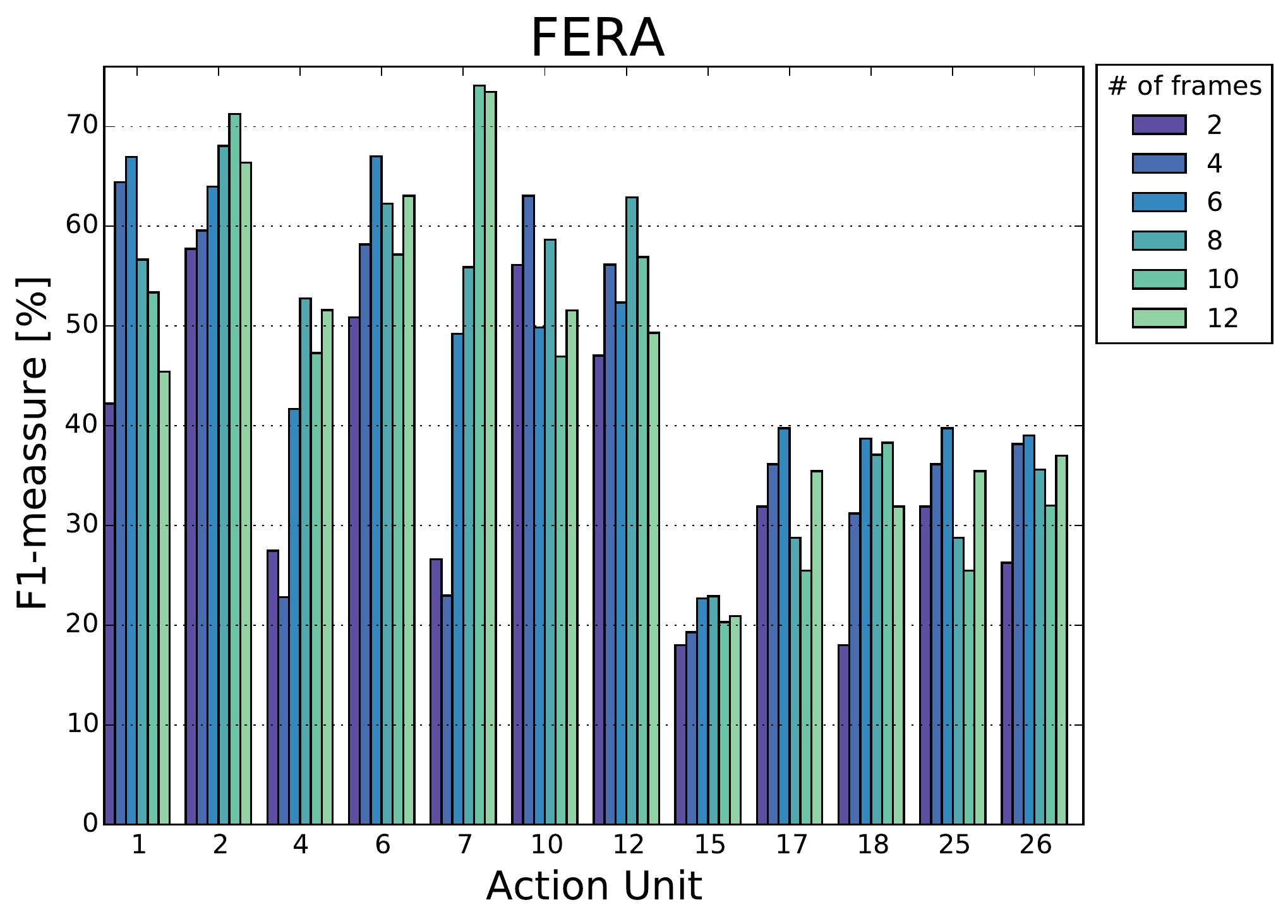}
\end{subfigure}%
\caption{F1-meassure per AU for different window sizes for the frame-based VSLem detection.}
\label{fig:cv_frame}
\end{figure*}

We also compared the variable state models with recent methods for frame-based AU detection.
The first related method, Early Fusion (EF) \cite{EF}, applies a hierarchical Gaussianization and scale-invariant feature transform on motion features. The classification is done by SVMs.
In MKL \cite{MKL}, a kernalized SVM is trained for each AU and the outputs are averaged in order to exploit temporal information.
CoT (Cascade of Tasks) \cite{ding2013facial} is trained on sequences and applies segment-based detection. This approach is a combination of three simple algorithms for static-frame-level-detection, segment-level-detection and transition-level detection. 
Selective Transfer Machine (STM) \cite{chu2013selective} is based on static SVMs, which personalizes the generic SVM classifier by learning the classifier and re-weighting the training samples that are most relevant to the test subject during inference.
HMTMKL \cite{zhu2014multiple} is a method for multiple AU recognition. A multi-task feature learning (MTFL) algorithm is adopted to learn the shared features among AUs and recognize AUs simultaneously. The AU relations are then modeled by a Bayesian graphical model. Finally, \cite{zhang2014p} is also a multi task learning approach and applies simultaneous detection of multiple facial AUs by exploiting their inter-relationships. 

\begin{figure*}[th]
    \footnotesize
    \TopFloatBoxes
    \begin{floatrow}
        \capbtabbox{
            \rowcolors{3}{white}{tablegray}
\begin{tabularx}{\linewidth}{c|l|XXXXXX}
    \rowcolor{tablegray}
\toprule
      & AU  & SVM  & HCRF & HCORF & VSLm & VSLd & VSLem \\
      &     & (SB) &      &       &      &      &      \\
\midrule  
& 1   & 56.1 & 51.4 & 58.3  & 68.9 & 72.3 & \textbf{73.7} \\
& 2   & 60.9 & 67.3 & 68.0  & 71.5 & \textbf{77.4} & 76.3 \\
Upper & 4   & 61.8 & 63.0 & 57.3  & 68.4 & \textbf{72.3} & 66.4 \\
Face  & 5   & 51.3 & 73.1 & 76.9  & 75.2 & 77.2 & \textbf{81.3} \\
      & 6   & 68.8 & 70.5 & 64.2  & 74.3 & 72.2 & \textbf{74.8} \\
      & 9   & 71.4 & 70.3 & 67.7  & 68.5 & \textbf{73.5} & 72.2 \\
\midrule  
& 12  & 67.2 & 65.9 & 66.3  & \textbf{71.9} & 68.3 & 69.9 \\
& 15  & 52.7 & 61.3 & 56.4  & 64.4 & \textbf{68.7} & 68.5 \\
Lower & 17  & 60.5 & 62.4 & 55.3  & 61.2 & 73.4 & \textbf{74.3} \\
Face  & 20  & 57.3 & 61.5 & 57.2  & 63.4 & 71.8 & \textbf{73.2} \\
      & 25  & 63.8 & 71.2 & 68.4  & \textbf{74.2} & 72.3 & 72.4 \\
      & 26  & 63.5 & 64.4 & 64.8  & 67.3 & 64.2 & \textbf{68.4} \\
\midrule  
& Avg & 61.3 & 65.2 & 62.6  & 69.1 & 72.0 & \textbf{72.6} \\
\bottomrule
\end{tabularx}

        }
        {
        \caption{F1-sequence-based results on the \textbf{DISFA} database}
        \label{tab:disfa_sb}
        }
        \capbtabbox{
            \rowcolors{3}{white}{tablegray}
\begin{tabularx}{\linewidth}{c|l|llXXX}
    \rowcolor{tablegray}
\toprule
      & AU  & SVM           & VSLem               & HMTMKL                 & $l_p$MTMKL        & MTFL                  \\
      &     & (FB)          &                    & \cite{zhu2014multiple} & \cite{zhang2014p} & \cite{zhang2013multi} \\
\midrule                                                                                       
      & 1   & 53.5          & \textbf{75.8} (10) & 72                     & 74                & 61                    \\
      & 2   & 66.8          & 66.2 (6)           & 63                     & 64                & \textbf{70}           \\
Upper & 4   & 59.2          & 52.5 (12)          & 67                     & 68                & \textbf{76}           \\
Face  & 5   & \textbf{71.8} & 51.7 (6)           & 55                     & --                & ---                   \\
      & 6   & 58.8          & 65.3 (12)          & 70                     & \textbf{71}       & 65                    \\
      & 9   & 65.5          & \textbf{65.4} (12) & 63                     & --                & ---                   \\
\midrule                                                                                       
      & 12  & 63.8          & 68.6 (10)          & 72                     & \textbf{76}       & ---                   \\
      & 15  & 58.3          & \textbf{79.6} (10) & 69                     & 72                & 68                    \\
Lower & 17  & 55.9          & \textbf{79.0} (8)  & 60                     & 63                & 74                    \\
Face  & 20  & 58.3          & 69.5 (8)           & 68                     & 69                & \textbf{71}           \\
      & 25  & 62.6          & 65.5 (6)           & \textbf{79}            & 74                & ---                   \\
      & 26  & 68.7          & \textbf{71.5} (10) & 63                     & ---               & ---                   \\
\midrule                 
      & Avg & 61.9          & 67.6 (*69.1)       & 66.8                   & \textbf{70.1}     & 69.3                  \\
\bottomrule
\end{tabularx}

        }
        {
        \caption{F1-frame-based results on the \textbf{DISFA} database}
        \label{tab:disfa_fb}
        }
    \end{floatrow}
    \vspace{18pt}
    \begin{floatrow}
        \capbtabbox{
            \rowcolors{4}{white}{tablegray}
\begin{tabularx}{\linewidth}{c|l|XXXXXX}
\rowcolor{tablegray}
\toprule
      & AU  & SVM  & HCRF & HCORF & VSLm          & VSLd          & VSLem         \\
      &     & (SB) &      &       &               &               &               \\
\midrule      
      & 1   & 63.1 & 57.1 & 63.4  & \textbf{67.3} & 55.8          & 65.1          \\
      & 2   & 62.2 & 65.8 & 64.8  & 63.8          & 64.4          & \textbf{71.7} \\
Upper & 4   & 44.7 & 44.4 & 44.2  & 44.2          & \textbf{49.7} & 48.2          \\
Face  & 6   & 57.4 & 53.5 & 51.8  & \textbf{58.4} & 53.7          & 54.9          \\
      & 7   & 60.3 & 64.2 & 65.4  & 63.2          & 66.2          & \textbf{67.5} \\
      & 10  & 50.8 & 55.5 & 56.4  & \textbf{58.5} & 57.4          & 56.3          \\
\midrule      
      & 12  & 54.3 & 45.2 & 43.2  & 53.3          & \textbf{54.7} & \textbf{54.7} \\
      & 15  & 12.4 & 15.3 & 14.9  & 14.4          & 14.2          & \textbf{15.5} \\
Lower & 17  & 44.9 & 64.8 & 68.3  & 67.8          & 69.4          & \textbf{71.6} \\
Face  & 18  & 44.0 & 43.1 & 41.7  & \textbf{50.3} & 50.1          & 49.8          \\
      & 25  & 52.5 & 54.3 & 51.2  & \textbf{61.1} & 54.8          & 57.5          \\
      & 26  & 48.3 & 33.4 & 35.8  & \textbf{49.4} & 44.4          & 48.4          \\
\midrule    
      & Avg & 52.0 & 53.5 & 53.4  & 57.7          & 57.2          & \textbf{59.0} \\
\bottomrule
\end{tabularx}

        }
        {
        \caption{F1-sequence-based results on the \textbf{GEMEP-FERA} database}
        \label{tab:fera_sb}
        }
        \capbtabbox{
            \rowcolors{4}{white}{tablegray}
\begin{tabularx}{\linewidth}{c|l|llXXXXXXX}
\rowcolor{tablegray}
\toprule
            & AU  & SVM           & VSLem                  & CLM                   & CoT                   & STM                      & MKL        & EF        \\
            &     & (FB)          &                       & \cite{chew2011person} & \cite{ding2013facial} & \cite{chu2013selective}  & \cite{MKL} & \cite{EF} \\
\midrule                                                                                                                                                  
            & 1   & 52.5          & 66.9  (6)             & \textbf{78}           & 64.2                  & 68.1                     & 61.1       & 57.6      \\
            & 2   & 51.8          & 71.2  (10)            & \textbf{72}           & 57.2                  & 65.5                     & 54.4       & 49.4      \\
Upper       & 4   & 42.5          & \textbf{52.7}  (6)    & 43                    & 46.6                  & 43.3                     & 45.4       & 43.6      \\
Face        & 6   & 55.2          & 67.0  (6)             & 66                    & \textbf{72.9}         & 71.6                     & 67.0       & 62.3      \\
            & 7   & 53.3          & \textbf{74.1}  (10)   & 55                    & 67.4                  & 66.2                     & 65.1       & 61.3      \\
            & 10  & 44.9          & 63.0  (4)             & 47                    & ---                   & ---                      & ---        & ---       \\
\midrule                                                                                                                                                  
            & 12  & 42.2          & 62.9  (10)            & 78                    & \textbf{78.3}         & 82.1                     & 75.4       & 71.5      \\
            & 15  & 12.2          & 22.9  (10)            & 16                    & \textbf{39.3}         & ---                      & ---        & ---       \\
Lower       & 17  & 31.9          & \textbf{50.1}  (10)   & 47                    & 38.6                  & 35.9                     & 36.7       & 30.1      \\
Face        & 18  & 42.4          & 38.7  (6)             & \textbf{45}           & ---                   & ---                      & ---        & ---       \\
            & 25  & \textbf{41.3} & 39.7  (6)             & 31                    & ---                   & ---                      & ---        & ---       \\
            & 26  & 49.5          & 39.0  (6)             & \textbf{54}           & ---                   & ---                      & ---        & ---       \\
\midrule                                                                                                                                      
            & Avg & 46.5          & 58.7 (*\textbf{63.6}) & 56.0                  & 57.1                  & \textbf{61.8}            & 57.9       & 57.6    \\
\bottomrule
\end{tabularx}

        }
        {
        \caption{F1-frame-based results on the \textbf{GEMEP-FERA} database}
        \label{tab:fera_fb}
        }
    \end{floatrow}
\end{figure*}

\subsubsection*{Frame-based Results}

The experiments for per-frame AU detection were performed on the GEMEP-FERA and DISFA database, where we applied a sliding window to each frame in order to obtain the predictions per frame (by assigning the classifier's prediction to the central frame in the window).
For each AU, we cross-validated over different window sizes to find the optimal size per AU. The results are shown in Fig. \ref{fig:cv_frame}.
Interestingly, the average window size on the AUs from the GEMEP-FERA dataset is shorter than that of the AUs from the DISFA dataset.
This is mainly because both datasets contain facial expressions recorded in different contexts (acted vs. spontaneous), so this difference in the duration of the AU activations is expected. Also, in DISFA, the expressions are less dynamic because the participants respond spontaneously to the watched youtube videos, while in GEMEP-FERA, the participants are actors and show much more dynamic emotions like 'Anger' or 'Fear' with fast facial muscle movements. 

\begin{figure*}[t!]
    \centering
    \begin{subfigure}[t]{0.5\textwidth}
        \centering
        \includegraphics[height=0.7\linewidth]{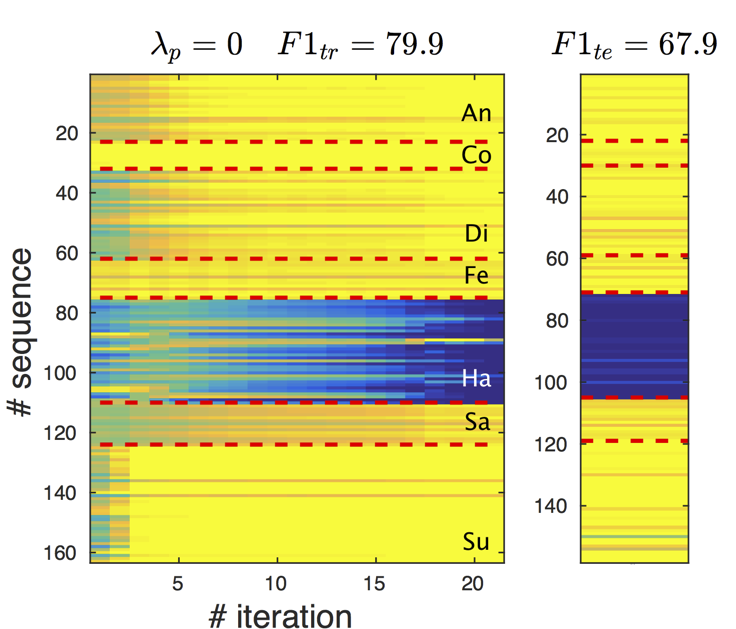}
        \caption{Posterior regularization not active}
    \end{subfigure}%
    \begin{subfigure}[t]{0.5\textwidth}
        \centering
        \includegraphics[height=0.7\linewidth]{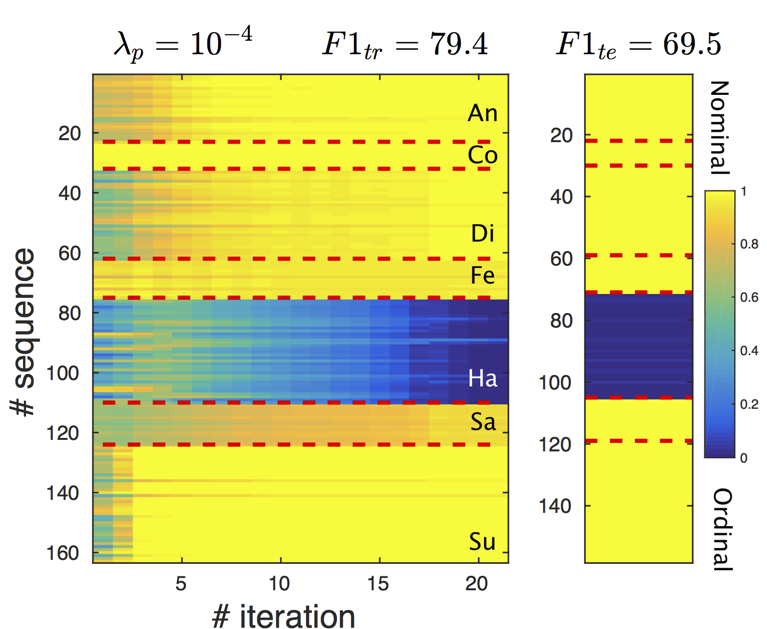}
        \caption{Posterior regularization active}
    \end{subfigure}%
         \caption{
        The visualization of the learning of the latent variable $\nu=\{ordinal,nominal\}$ within the VSLem model aplied to the CK+ database. The evaluation was performed using 2-fold subject-independent evaluation using equal number of training/test data. The posterior probabilities  $P(\nu|y,x)$  are shown for each sequence after the maximization step of each EM-iteration.
The plots in (a) and (b) shows the learning process without and with the posterior regularization (using the optimal valdiation parameter $\lambda_p$).
    }
    \label{fig:posterior}
\end{figure*}


Table \ref{tab:fera_fb} shows the F1-measure for the detection of each AU from the GEMEP-FERA dataset with the window size reported in brackets.
The STM \cite{chu2013selective}, despite the subject adaptation, still fails to reach the full performance of the VSLem model on the mutual set of evaluated AUs. 
This is attributed to the fact that the STM does not model the temporal dynamics. But again, different settings were used in these evaluations. These results demonstrate again that the assignment of both types of latent states, as done in the VSL-CRF models is critical for achieving superior performance on this task. Table \ref{tab:disfa_fb} shows the results on the DISFA dataset.
The two Multi-task learning approaches (MTL) \cite{zhang2014p, zhang2013multi} apply simultaneous detection of multiple facial AUs by exploiting their inter-relationships. They also model the correlation among AUs which results in the very high detection rate. The proposed VSL-CRF model reaches the results that are comparable with that of the state-of-the-art. The high F1-frame-based score achieved by both methods demonstrates the importance of both the modeling of  the inter-relationships of AUs, as done in the former, and dynamics, as done in the latter.
\begin{figure*}[t]
    \footnotesize
    \TopFloatBoxes
    \begin{floatrow}
        \capbtabbox{
            \rowcolors{3}{white}{tablegray}
            \begin{tabularx}{\linewidth}{c|XXXXXX}
                \toprule
                AU  & SVM           & HCRF          & HCORF         & VSLm        & VSLd        & VSLem         \\
                \midrule
                1   & 40.0          & \textbf{44.2} & 39.7          & 42.6          & 43.3          & 43.7          \\
                2   & 40.4          & 44.8          & \textbf{47.4} & 43.3          & 42.8          & 41.2          \\
                4   & 33.3          & 33.5          & 25.3          & 22.4          & \textbf{34.8} & 34.0          \\
                6   & 57.7          & 54.1          & 46.3          & \textbf{58.7} & 49.7          & 54.5          \\
                12  & 23.7          & 35.9          & 34.5          & 33.0          & 36.0          & \textbf{37.4} \\
                17  & 22.2          & 29.4          & 16.9          & 19.8          & 24.6          & \textbf{25.6} \\
                25  & 37.2          & \textbf{67.9} & 44.6          & 46.7          & 44.4          & 45.6          \\
                26  & \textbf{37.5} & 31.4          & 36.0          & 37.3          & 33.2          & 32.4          \\
                \midrule
                AVG & 35.3          & 38.9          & 36.3          & 37.9          & 38.6          & \textbf{39.2} \\
                \bottomrule
            \end{tabularx}
        }
        {
\caption{Per-sequence classification rate on the cross dataset experiment  DISFA $\rightarrow$ \textbf{GEMEP-FERA}}%
            \label{tab:CD_D_F}
        }
        \capbtabbox{
            \rowcolors{3}{white}{tablegray}
            \begin{tabularx}{\linewidth}{c|XXXXXX}
                \toprule
                AU  & SVM           & HCRF          & HCORF         & VSLm        & VSLd        & VSLem         \\
                \midrule
                1   & 28.5          & 32.8          & 29.6          & 35.2          & \textbf{40.0} & 35.2          \\
                2   & 37.2          & 45.7          & 41.3          & \textbf{49.4} & 49.3          & 40.9          \\
                4   & 25.9          & 44.9          & 29.2          & 24.8          & 35.5          & \textbf{40.9} \\
                6   & \textbf{50.8} & 44.6          & 39.9          & 48.6          & 42.6          & 48.4          \\
                12  & 21.2          & 32.1          & 26.2          & \textbf{42.7} & 28.5          & 39.6          \\
                17  & 26.6          & 23.1          & 21.9          & 25.4          & 25.3          & \textbf{32.7} \\
                25  & 42.1          & 46.5          & 50.5          & 52.6          & \textbf{53.2} & 45.3          \\
                26  & 22.2          & 34.1          & 33.0          & 33.3          & 37.8          & \textbf{38.0} \\
                \midrule
                AVG & 34.3          & 36.7          & 33.9          & 39.0          & 39.1          & \textbf{40.1} \\
                \bottomrule
            \end{tabularx}
        }
        {
\caption{Per-sequence classification rate on the cross dataset experiment  GEMEP-FERA $\rightarrow$ \textbf{DISFA}}%
            \label{tab:CD_F_D}
        }
    \end{floatrow}
\end{figure*}

\subsubsection{Sequence-based Cross-database Results}
Detecting AUs across datasets is challenging because of differences in contexts in which this data is recorded (acted vs. spontaneous, illumination, frame rate, etc.). In this experiment, we apply the VSL-CRF models, the H-CRF and H-CORF models, and the baseline SVM on the pre-segmented sequence form the AU databases GEMEP-FERA and DISFA. Table \ref{tab:CD_F_D}  and \ref{tab:CD_D_F} show the results for the experiment in which we trained the models using the GEMEP-FERA database and evaluated them on the DISFA database, and the other way round, respectively. We observe that in this setting also the proposed VSL-CRF models outperform nominal- or ordinal-state methods, and the static SVM. This demonstrates the strong generalization capability of the proposed models. It is interesting to note that this difference is much smaller in the results reported in Table \ref{tab:CD_F_D}, where HCRF achieves similar results to VSLem, compared to Table \ref{tab:CD_D_F}, where the HCRF and H-CORF are largely outperformed by the  VSL-CRF models. We attribute this to the fact that the acted data (GEMEP-FERA) contains much more variation in facial expressions compared to spontaneous expressions in DISFA dataset. Consequently, the models are learned on more diverse data, allowing them to generalize better to subtle facial expressions, as evidenced by this experiment. We also observe that all three VSL-CRF learning approaches perform similarly in this setting. A possible reason is that since the data distributions vary significantly across the datasets (in terms of number of active examples, as well as the AU co-occurrences), this limits the proposed learning approaches to reach their full performance. Finally, note that the perfomance on the both datasets drops significantly compared to the results in Tables \ref{tab:disfa_sb} and \ref{tab:fera_sb}. For exmaple, for GEMEP-FERA, the results on the used set of AUs from from 60.2$\%$ to 39.2$\%$ for the best performing model. This indicates the importance of accounting for the dataset-differences during modeling of facial expressions. 

\subsubsection{The Effect of Posterior Regularization}
On all datasets, the VSLd and VSLem outperforms VSLm. This is mainly attributed to the more flexible representation of the latent states as well as the additional posterior regularization. To get some insights into the behavior of the posterior regularization during the learning process, we performed additional experiment on the CK+ dataset. Specifically, we trained the VSLem model with and without the posterior regularization and monitored the parameter for each EM-iteration (the graphs showing the changes in the nominal/ordinal states on the training data). The training/test sets consisted of 162 sequences each, and are sorted according to the sequence label. The results are shown in Fig. \ref{fig:posterior}. The bar on the right side of each main figure shows the contribution of ordinal/nominal states for the prediction of the test sequences. We can see that the emotion happiness exhibited a strong ordinal structure as encoded with its ordinal states, while the other emotion were predicted using the nominal states. The figure on the right shows the same learning process with active posterior regularization. Again, the emotion happiness was trained and predicted using mainly the ordinal states but all other emotions mainly preferred using the nominal states during training and inference, as the result of the regularization. The learned type of the latent states is also consistent on the test data.  Finally, although only emotion happiness showed strong ordinal nature, as learned from the employed features of facial expressions, the nominal states selected for the other emotion categories do not imply that there is no ordinal structure in their facial expressions but that the nominal states were a better fit for the target data used in this experiment. Note also that when the posterior regularization is used, the F1-sequence-based measure on the test sets is higher (69.5$\%$ vs. $67.9\%$), demonstrating the benefit of the posterior regularization. Furthermore, note that this regularization enforces the model to converge to either nominal or ordinal states during the model learning. 
 
\section{Conclusions}
\label{sec:conc}
In this paper, we proposed a novel Variable-State Conditional Random Field model for dynamic facial expression recognition and AU detection. 
By allowing the structure of the latent states of target classes to vary for each target sequence, the proposed model can better discriminate between different facial expressions than the existing models that restrict their latent states to have the same and pre-defined structure for all classes (nominal or ordinal). For this model, we proposed two novel learning strategies and the posterior regularization of the latent states, resulting in a more robust model for the target tasks. This leads to superior performance compared to traditional latent CRF models. We also showed on three facial expression datasets that the proposed model performs similar or better than the state-of-the-art for the task of sequence-based facial expression recognition, and that it reaches state-of-the-art performance for the task of per-frame AU-detection. The future work should focus on more detailed analysis of the learning of the target latent states within each emotion class and AU (e.g., the automated selection of the window size for each AU), as well as analysis of the relations between the learned latent states and the temporal aspects of facial expressions such as their temporal phases and intensity. Also, extending the proposed approach so that it can handle simultaneous detection of multiple AUs, and its adaptation to previously unseen datasets, are also interesting avenues to pursue. 
\section*{Acknowledgments}
This work has been funded by the European Community Horizon 2020
[H2020/2014-2020] under grant agreement no. 645094 (SEWA). The work of Vladimir Pavlovic has been funded by the National Science Foundation under Grant no. IIS0916812.

\bibliographystyle{unsrt}
\footnotesize
\bibliography{bibfile}

\newpage
\begin{figure}[H]

  \begin{minipage}[c]{0.3\textwidth}

    \includegraphics[width=\textwidth]{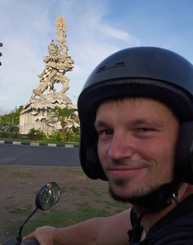}
  \end{minipage}\hfill
  \begin{minipage}[c]{0.67\textwidth}
      \footnotesize
    \textbf{Robert Walecki} received his MSc degree in Physics from the The Ruprecht-Karls-University in Heidelberg, Germany, in 2013.
    He is currently working towards his Ph.D. degree at the Department of Computing,
    Imperial College London, London, UK.
    His research interests span the areas of Computer Vision, Pattern Recognition, Machine Learning and, in particular, human-computer interaction and automatic human behavior analysis.
  \end{minipage}

\vspace{1cm}

  \begin{minipage}[c]{0.3\textwidth}
    \includegraphics[width=\textwidth]{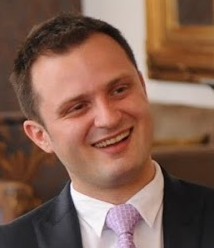}
  \end{minipage}\hfill
  \begin{minipage}[c]{0.67\textwidth}
      \footnotesize
    \textbf{Ognjen Rudovic} received his BSc degree in Automatic Control from Faculty of Electrical Engineering, University of Belgrade, Serbia, in 2007, MSc degree in Computer Vision from Computer Vision Center (CVC), Universitat Autonoma de Barcelona, Spain, in 2008, and PhD in Computer Science from Imperial College London, UK. He is currently a Research Fellow at the Computing Department, Imperial College London, UK. His research interests are in automatic recognition of human affect, machine learning and computer vision.
  \end{minipage}

\vspace{1cm}

  \begin{minipage}[c]{0.3\textwidth}
    \includegraphics[width=\textwidth]{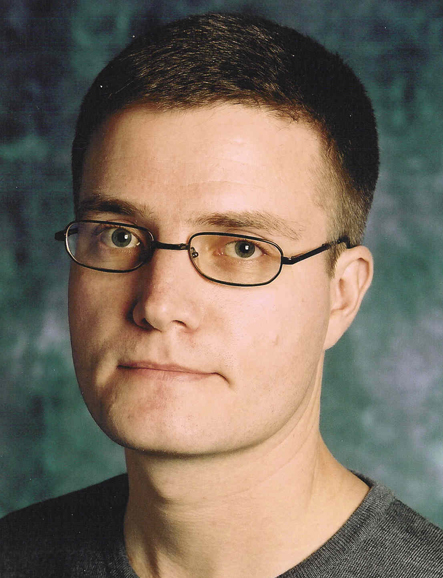}
  \end{minipage}\hfill
  \begin{minipage}[c]{0.67\textwidth}
      \footnotesize
\textbf{Vladimir Pavlovic} 
received the PhD degree in
electrical engineering from the University of
Illinois at Urbana-Champaign in 1999. From
1999 until 2001, he was a member of the
research staff at the Cambridge Research
Laboratory, Massachusetts. He is
an associate professor in the Computer Science
Department at Rutgers University, New Jersey. Before joining Rutgers in 2002, he
held a research professor position in the
Bioinformatics Program at Boston University. His research interests
include probabilistic system modeling, time-series analysis,
computer vision, and bioinformatics.
  \end{minipage}

\vspace{1cm}
  \begin{minipage}[c]{0.3\textwidth}
    \includegraphics[width=\textwidth]{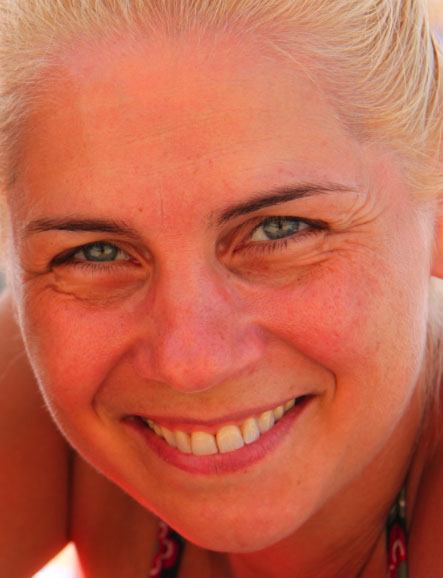}
  \end{minipage}\hfill
  \begin{minipage}[c]{0.67\textwidth}
      \footnotesize
      \textbf{Maja Pantic} is Professor in
Affective and Behavioural Computing at Imperial
College London, Computing Dept., UK,
and at the University of Twente, Dept.
of Computer Science, Netherlands. She received
various awards for her work on automatic
analysis of human behaviour including the European
Research Council Starting Grant Fellowship
2008 and the Roger Needham Award 2011.
She currently serves as the Editor in Chief of
Image and Vision Computing Journal, and as an
Associate Editor for IEEE Trans. on Systems, Man, and
Cybernetics Part B and IEEE TPAMI.
  \end{minipage}

\end{figure}

\end{document}